%% file: paper.tex
\documentclass[]{bytedance_seed}

\usepackage[toc,page,header]{appendix}

\usepackage{wrapfig}
\usepackage{xcolor}
\usepackage{minitoc}

\usepackage{booktabs}
\usepackage{tabularx}
\usepackage{makecell}
\usepackage{tikz}
\usepackage{enumitem}
\tcbuselibrary{skins}

\usetikzlibrary{shapes.geometric, arrows.meta, positioning, fit, shadows}

\title{WideSearch: Benchmarking Agentic Broad Info-Seeking}

\author[*]{Ryan Wong}
\author[*]{Jiawei Wang}
\author[]{Junjie Zhao}
\author[]{Li Chen}
\author[]{Yan Gao}
\author[]{Long Zhang}
\author[]{Xuan Zhou}
\author[]{Zuo Wang}
\author[]{Kai Xiang}
\author[]{Ge Zhang}
\author[]{Wenhao Huang}
\author[\dagger]{Yang Wang}
\author[\dagger]{Ke Wang}

\affiliation[]{ByteDance Seed}

\contribution[*]{Co-first authors}
\contribution[\dagger]{Corresponding authors}

\abstract{
From professional research to everyday planning, many tasks are bottlenecked by wide-scale information seeking, which is more repetitive than cognitively complex. With the rapid development of Large Language Models (LLMs), automated search agents powered by LLMs offer a promising solution to liberate humans from this tedious work. However, the capability of these agents to perform such "wide-context" collection reliably and completely remains largely unevaluated due to a lack of suitable benchmarks. To bridge this gap, we introduce WideSearch, a new benchmark engineered to evaluate agent reliability on these large-scale collection tasks. The benchmark features 200 manually curated questions (100 in English, 100 in Chinese) from over 15 diverse domains, grounded in real user queries. Each task requires agents to collect large-scale atomic information, which could be verified one by one objectively, and arrange it into a well-organized output. A rigorous five-stage quality control pipeline ensures the difficulty, completeness, and verifiability of the dataset. We benchmark over 10 state-of-the-art agentic search systems, including single-agent, multi-agent frameworks, and end-to-end commercial systems. Most systems achieve overall success rates near 0\%, with the best performer reaching just 5\%. However, given sufficient time, cross-validation by multiple human testers can achieve a near 100\% success rate. These results demonstrate that present search agents have critical deficiencies in large-scale information seeking, underscoring urgent areas for future research and development in agentic search.

}

\date{\today}
\correspondence{Yang Wang at \email{wangyang.127@bytedance.com}, Ke Wang at \email{wangke@bytedance.com}}

\checkdata[Project Page]{\url{https://widesearch-seed.github.io/}}

\begin{document}
\maketitle


\input{sections/intro_update}
\input{sections/relatedwork}
\input{sections/benchmark}
\input{sections/approach}
\input{sections/experiments}

\input{sections/conclusion}
\clearpage
\input{sections/acknowledgement}

\clearpage

\bibliographystyle{plainnat}
\bibliography{main}

\clearpage

\beginappendix

\input{sections/appendix}

\end{document}

%% file: sections/intro_update.tex
\section{Introduction}

With the advent of advanced agentic frameworks such as OpenAI DeepResearch \cite{OpenAI_DRSC} and Manus \cite{Manus2025}, the development of agent systems based on Large Language Models (LLMs) is entering its second half, where the focus is rapidly shifting from demonstrating novel capabilities to achieving practical, real-world reliability. This transition is driven by a fundamental recognition of the inherent limitations in standalone models: their finite parameters make it impossible to store all knowledge, the prohibitive cost of retraining makes them lag behind real-time information, and they naturally struggle with long-tail or specialized facts. Consequently, in this evolving domain, the ability to effectively utilize search tools has become paramount. The most critical question in this race is no longer just what an agent can do, but how we can measure and improve its ability to leverage search in authentic user scenarios to deliver tangible value and drive meaningful product iteration.

\begin{figure}[t]
    \centering
    \includegraphics[width=\linewidth]{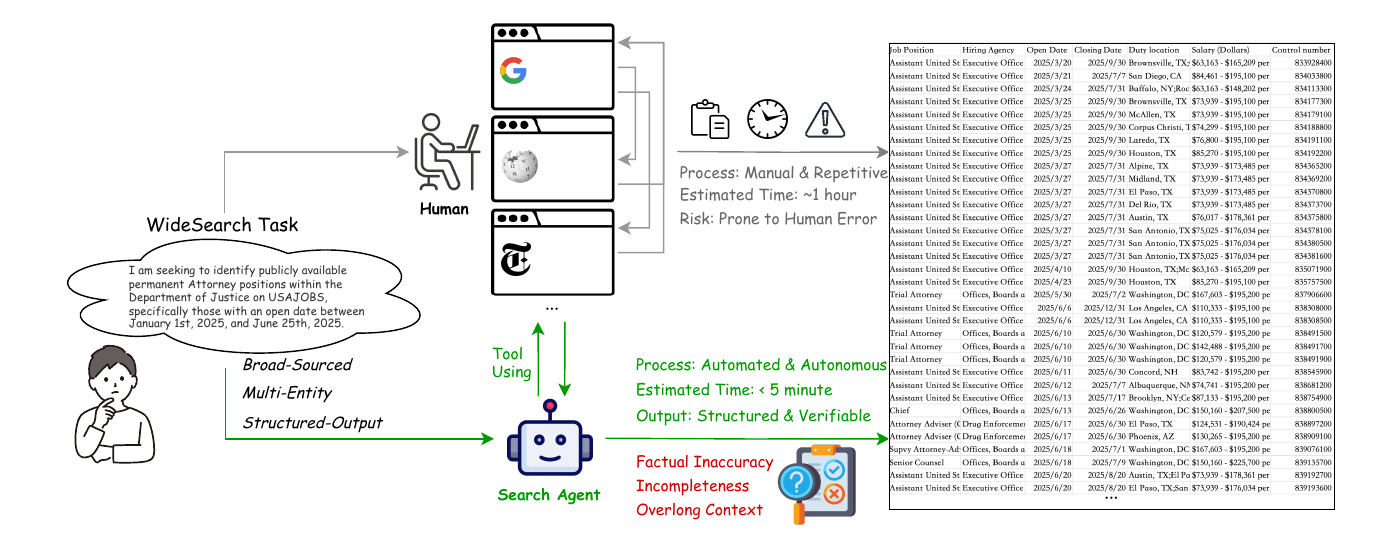}
    \caption{A conceptual comparison of manual and agent-based approaches for WideSearch tasks. The diagram illustrates the operational workflow and inherent limitations associated with two distinct methodologies for large-scale information seeking. It contrasts the labor-intensive nature of the traditional manual approach with the potential efficiencies and novel failure modes of automated search agents. This comparison underscores the necessity for a systematic evaluation to quantify agent performance and reliability.}
    \label{fig:overview}
\end{figure}

\begin{figure}[t]
    \centering
    \includegraphics[width=\linewidth]{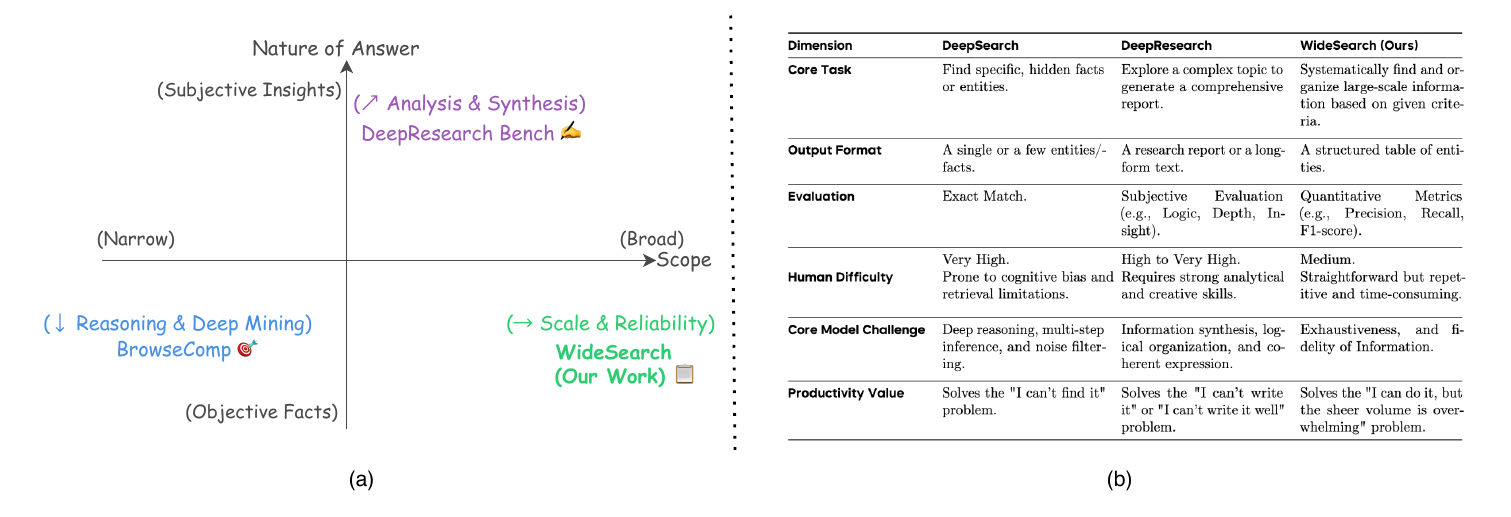}
    \caption{An overview and detailed comparison of DeepSearch, DeepResearch, and our WideSearch. The conceptual map on the left (a) illustrates the high-level relationships and operational domains of the three paradigms. The table on the right (b) provides a detailed breakdown, contrasting them across key dimensions including core tasks, evaluation methods, and primary value propositions.}
    \label{fig:intro}
\end{figure}

Our in-depth analysis of real-world user queries reveals a significant gap: a common and critical class of information-seeking tasks is not adequately evaluated by existing agent benchmarks. We term this category WideSearch, which involves tasks that require an agent to thoroughly and accurately acquire all large-scale atomic information meeting a series of criteria, and then arrange it in a well-organized output. For example, a financial analyst may need to find all companies in a sector that meet specific revenue and growth criteria, or a job seeker may need to find all job vacancies that match their criteria for role, location, and experience level. For humans, executing such tasks is excruciatingly tedious; as depicted in Figure \ref{fig:overview}, the transition from this laborious manual process to an automated agent workflow promises immense efficiency gains, but also introduces new failure modes that demand systematic evaluation. Consequently, WideSearch carves out a distinct problem space. As illustrated in Figure \ref{fig:intro}, it diverges from DeepSearch, which targets the "I can't find it" problem of locating specific, hard-to-find facts, and DeepResearch, which addresses the "I can't write it well" problem of synthesizing complex narratives. Instead, WideSearch tackles tasks whose primary challenge is not cognitive difficulty but operational scale and fidelity—the "I could do it, but the sheer volume is overwhelming" problem—a domain largely overlooked by current benchmarks.

To systematically evaluate this paradigm, we introduce WideSearch, the first benchmark specifically designed for this purpose, supported by a sophisticated multi-stage data collection and verification framework, as well as a hybrid automated evaluation system that ensures objectivity. Benchmarking more than 10 state-of-the-art agent systems reveals a stark reality: current systems are profoundly challenged by the demands of comprehensiveness and fidelity at scale. The overall success rate is exceptionally low, with even the top-performing multi-agent framework achieving a mere 5.1\%, and individual humans also struggling at 20\%. Our key insight, derived from a test-time scaling analysis, is that this failure does not stem from an inability to find individual facts—item-level F1 scores can approach 80\% with sufficient retries. Rather, the bottleneck is that we must ensure the absolute completeness and accuracy of each atomic unit of information within a large-scale search. Any single data omission or error, or the integration of extra data into the final result, results in total failure of the task execution. Our detailed error analysis traces this failure to fundamental deficiencies in advanced agentic capabilities, such as incomplete planning, a lack of reflection to iterate on failed searches, and the inability to correctly use retrieved evidence. These findings provide a clear roadmap, suggesting that future progress hinges on developing more sophisticated agent models and architectures, particularly multi-agent frameworks that enable parallel search and cross-validation, mimicking the collaborative human processes required to tackle these complex, large-scale tasks.

%% file: sections/relatedwork.tex
\section{Related Work}

\subsection{Benchmarks for Search Agents}

The evaluation of search agents has evolved significantly, moving from simple fact retrieval to complex, multi-step reasoning tasks \cite{lei2025spider}. Early benchmarks such as Natural Questions \cite{kwiatkowski2019natural} and TriviaQA \cite{joshi2017triviaqa} established a foundation for question answering, but often tested information that could be retrieved with a single query or was already contained within a model's parametric knowledge. The subsequent development of multi-hop QA datasets, including HotpotQA \cite{yang2018hotpotqa}, 2WikiMultiHopQA \cite{ho2020constructing}, and Musique \cite{trivedi2022musique}, increased the complexity by requiring agents to connect multiple pieces of evidence to derive an answer. However, these tasks typically feature a structured, linear path to the solution and do not fully capture the ambiguity and non-linear exploration required in real-world search scenarios.

More recent benchmarks have embraced this complexity, focusing on what we categorize as "DeepSearch": intensive, vertical investigations into a single, complex topic. For instance, GAIA \cite{mialon2023gaia} presents challenging multi-hop questions that push the boundaries of reasoning. Similarly, Xbench-DeepSearch \cite{chen2025xbench} specifically targets agents’ deep search and tool-use capabilities through professionally annotated, dynamic tasks. Benchmarks like BrowseComp-en/zh \cite{wei2025browsecomp,zhou2025browsecomp} further elevate the difficulty by designing tasks with intricately coupled entities and deliberate information obfuscation, demanding sophisticated, non-linear exploration to reduce a high degree of initial uncertainty. Concurrently, the community has also explored the evaluation of comprehensive report generation. A notable example is the DeepResearch Bench \cite{du2025deepresearch}, which assesses an agent's ability to tackle PhD-level questions and synthesize the findings into a detailed, accurate report. Unlike existing benchmarks that test deep reasoning on a single query, WideSearch evaluates an agent's ability to gather broad information across multiple parallel entities by requiring it to populate a structured table.


\subsection{Search Agents}

The development of advanced Search Agents has been propelled by both proprietary and open-source efforts. Following initial breakthroughs from systems like OpenAI's Deep Research Agents \cite{OpenAI_DRSC} and Google's Gemini Deep Research \cite{GoogleGeminiDeepResearch2025}, a wave of related works emerged. Proprietary systems such as Grok-3 Deep Research \cite{xAIGrok3Beta2025} and Kimi-Researcher \cite{KimiResearcher2025} have demonstrated impressive, often superhuman, performance on complex information synthesis tasks. However, their closed-source nature and opaque training methodologies limit community-driven research and reproducibility.

In parallel, the open-source community has pursued two primary research directions. The first focuses on model-centric optimization, primarily through Reinforcement Learning (RL) to train agents end-to-end. Examples include R1-Searcher \cite{song2025r1} and Search-R1 \cite{jin2025search}, which train on local corpus, and DeepResearcher \cite{zheng2025deepresearcher}, which uses real search engines. To cut down on interaction costs, ZeroSearch \cite{sun2025zerosearch} trains an LLM to simulate a search engine, R1-Searcher++ \cite{song2025r1++} improves this by separating internal knowledge from external retrieval with a memory mechanism, and IKEA \cite{huang2025reinforced} utilizes the knowledge-boundary enhanced RL to reduce redundant retrieval.  Other efforts like WebDancer \cite{wu2025webdancer} and WebSailor \cite{li2025websailor} focus on generating high-quality synthetic data. The second direction is workflow and agent orchestration, which involves designing multi-agent systems. WebThinker \cite{li2025webthinker} uses specialized modules for problem-solving and report-writing, while Alita \cite{qiu2025alita} features a manager agent that can dynamically create MCP tools. However, the performance of these agents on broad information-seeking tasks hasn't been thoroughly evaluated. Our work, WideSearch, is the first benchmark specifically designed to assess search agents on this capability, paving the way for future development.

%% file: sections/benchmark.tex
\section{WideSearch Benchmark}

The construction of the WideSearch benchmark is a meticulous, human-centered process designed to ensure that each task is challenging, realistic, and aligned with our goal of evaluating wide-context information gathering. The entire workflow, from question design to final inclusion, is governed by a strict set of criteria and a multi-stage validation protocol.

\subsection{Task Definition}

The fundamental task in the WideSearch benchmark challenges an LLM agent to act as a diligent information seeker. Given a complex natural language query and a predefined table schema, the agent's objective is to populate the table by systematically gathering, synthesizing, and verifying information from the live web. This emulates real-world information-seeking scenarios that require discovery and aggregation rather than simple fact retrieval.

Formally, each task instance in WideSearch is defined by a tuple $(Q, S)$, where:
\begin{itemize}
    \item \textbf{A Query ($Q$):} A natural language question that implicitly specifies a set of target entities and the information required about them. For example, $Q$ could be: "I want to apply for full-time Master’s programs in civil engineering starting in 2026. Could you help me find the minimum GPA requirements for admission to Ivy League institutions in the US and Group of Eight universities in Australia?"
    \item \textbf{A Table Schema ($S$):} A predefined structure $S = \{C_1, C_2, \dots, C_m\}$, where each $C_j$ is a column header representing an attribute to be retrieved (e.g., `Country', `University', `Alliance', `Minimum GPA Requirement'). The schema defines the exact structure of the required output for objective evaluation.
\end{itemize}

The agent's goal is to interact with a web environment, primarily via search tools, to produce a final, populated table, $T_{\text{agent}}$. This objective decomposes into two primary challenges:

\begin{itemize}
    \item \textbf{Entity Set Identification:} The agent must first identify the complete and correct set of entities, $E = \{e_1, e_2, \dots, e_n\}$, that satisfy the constraints of the query $Q$. In this example, the entities are the 8 Ivy League institutions and the 8 Group of Eight universities. This tests the agent's ability to conduct a comprehensive search across different domains (US and Australian higher education).
    \item \textbf{Attribute Filling:} For each identified entity $e_i \in E$, the agent must find the corresponding values for each attribute $\{C_1, C_2, \dots, C_m\}$ defined in the schema $S$, sourcing the information from web pages.
\end{itemize}

The final output, $T_{\text{agent}}$, is therefore a table with $n$ rows (one for each identified entity) and $m$ columns (as defined by $S$), where each cell $T_{\text{agent}}(i, j)$ contains the value of attribute $C_j$ for entity $e_i$. The quality of this output is then measured against a ground-truth table to assess its completeness and factual accuracy.

\subsection{Task Construction Methodology}

The construction of tasks within the WideSearch benchmark is guided by a rigorous, principled methodology to ensure their quality, relevance, and alignment with the challenges of wide-context information seeking. Each task is manually curated by domain experts and must satisfy the following six fundamental principles:

\begin{itemize}
    \item \textbf{High Search Volume and Breadth}: Tasks are defined by their extensive informational breadth, requiring the agent to collate numerous distinct data points across multiple entities. This inherent breadth necessitates a high volume of search interactions and a prolonged, multi-step procedural trajectory for completion, distinguishing them from tasks that require only a singular, deep line of inquiry.
    \item \textbf{Temporal and Contextual Invariance}: The ground-truth answers exhibit high stability. They are static over time and are independent of geographical, ideological, or socio-cultural contexts, thereby guaranteeing the benchmark's long-term validity and global applicability.
    \item \textbf{Objective Verifiability}: Each task is associated with a deterministically verifiable set of facts. This allows for objective, consistent, and reproducible scoring against a predefined gold standard.
    \item \textbf{Public Accessibility}: The entire corpus of information required to formulate a complete answer is guaranteed to be publicly accessible via standard web search engines, ensuring that tasks are solvable without privileged access to information.
    \item \textbf{Reliance on External Tools}: Tasks are explicitly designed to exceed the bounds of an LLM's parametric knowledge. Successful completion is therefore contingent upon the agent's ability to engage in active, iterative, and effective web search, rather than relying on memorized information.

    \item \textbf{Scenario Diversity}: The benchmark encompasses a heterogeneous collection of scenarios spanning multiple distinct industries. This cross-domain diversity ensures that we evaluate generalizable agent capabilities, such as planning and synthesis, rather than task-specific or domain-dependent knowledge.
\end{itemize}

\begin{figure}[t]
    \centering
    \includegraphics[width=1.0\linewidth]{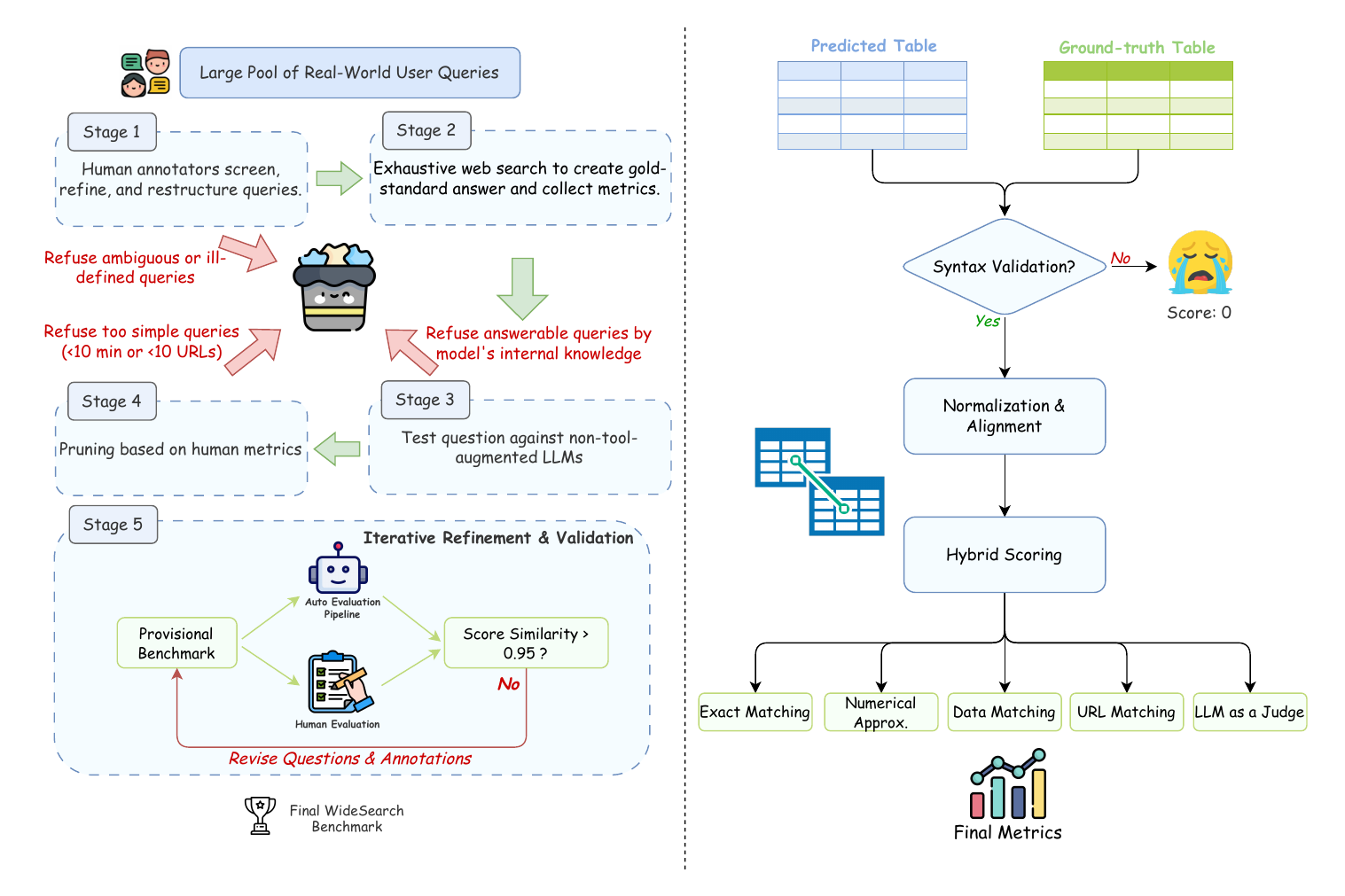}
    \caption{An overview of our integrated data pipeline, detailing the five-stage data curation and validation pipeline (left), and the automated evaluation pipeline (right).}
    \label{fig:curation_pipeline}
\end{figure}

\subsection{Data Curation and Validation Process}

To ensure every question in the benchmark rigorously adheres to the design principles, we implement a multi-stage data curation and validation pipeline, as illustrated in Figure \ref{fig:curation_pipeline}. This process transforms raw, real-world information needs into standardized, high-quality evaluation tasks. A task is only accepted into the final benchmark after successfully passing through all filtering stages and the final iterative validation loop.

\textbf{Sourcing and Refinement of Candidate Questions}: The process begins by sourcing a large pool of questions from real user queries, covering a wide array of domains such as finance, education, healthcare, and entertainment. These raw queries are often ill-defined or ambiguous. Human annotators meticulously screen these queries, selecting those with the potential to become good "wide-search" tasks. They then refine and restructure these selections into a clear, unambiguous candidate question set that aligns with our design principles.

\textbf{Gold Standard Annotation and Metric Collection}: Each candidate question is assigned to a human annotator. The annotator's task is to conduct an exhaustive web search to find and compile a comprehensive, gold-standard answer. During this process, they are required to meticulously record a set of key performance indicators: the total time to completion, the number of distinct search queries issued, the specific keywords used in each query, and the total number of unique web pages consulted to formulate the final answer.

\textbf{Parametric Knowledge Filtering}: To guarantee that tasks necessitate tool use, we subject each candidate question to a parametric knowledge test. The question is posed to a suite of powerful, non-tool-augmented LLMs. If any model can generate a complete and correct answer using only its internal knowledge, the question is discarded. This critical filtering step ensures that all tasks in WideSearch genuinely evaluate an agent's search and synthesis capabilities.

\textbf{Difficulty-Based Pruning}: We leverage the performance metrics collected by human annotators in previous steps to perform a quantitative difficulty assessment. Any task that does not meet our minimum complexity threshold is discarded. Based on our current heuristics, this includes any task that a human annotator completes in less than 10 minutes or by consulting fewer than 10 unique web pages.

\textbf{Iterative Refinement and Validation}: Tasks that pass the initial four-stage funnel form a provisional benchmark. This set then undergoes a final, iterative validation loop designed to align our automated scoring with human judgment. For each task, we first crawl a response through an existing commercial agentic system. Then, we use our automated evaluation system to rate this response. In parallel, we have human experts rate the same response. Then we compare the results from our automated evaluation pipeline with the results from expert human evaluators. If the evaluation results show a discrepancy (i.e., the similarity between automated and human scores is below our 95\% threshold), the task and its gold-standard annotation are flagged for revision. This cycle continues until the automated metrics reliably mirror human assessment, ensuring the integrity and reliability of the benchmark.

This rigorous, five-stage pipeline ensures that every task in the final WideSearch benchmark is grounded in a real-world need, demonstrably complex, verifiable, and resistant to simple memorization. Most critically, it also ensures that the automated evaluation for each task is calibrated to and predictive of human judgment. This final validation loop provides a strong guarantee that WideSearch is a robust and reliable testbed for advanced search agents. An illustrative example of a final task is provided in Figure \ref{fig:example_with_eval}.

\begin{figure}[h!]
    \centering
    
    \begin{tcolorbox}[
        enhanced,
        title=Task Prompt,
        colback=blue!5!white,
        colframe=blue!75!black,
        fonttitle=\bfseries,
        attach boxed title to top left={yshift=-2mm, xshift=3mm},
        boxed title style={
            colback=blue!75!black,
        }
    ]
    \small
    Could you list every single concert on Taylor Swift’s tour from January 1, 2010, to May 1, 2025, including the specific date, the concert’s English name, the country, the city, and the venue? Each show should be on its own line, in chronological order from earliest to latest. Please organize the results in one Markdown table with the following columns: Date, Concert’s English Name, Host Country, Host City, Host Venue
    
    Notes: Do not use date ranges for Date, list it in the format of “Day Month, Year”, for example: 4th June, 2011
    
    The output format is \verb|```markdown\n{data_content}\n```|.
    \end{tcolorbox}

    \vspace{0.5em}

    \begin{tcolorbox}[
        enhanced,
        title=Ground-Truth,
        colback=green!5!white,
        colframe=green!60!black,
        fonttitle=\bfseries,
        attach boxed title to top left={yshift=-2mm, xshift=3mm},
        boxed title style={
            colback=green!60!black,
        }
    ]
    \centering
    \begin{tabular}{@{} l l l l l @{}}
        \toprule
        \textbf{Date} & \textbf{Concert's English Name} & \textbf{Host Country} & \textbf{Host City} & \textbf{Host Venue} \\
        \midrule
        4th February, 2010 & Fearless Tour & Australia & Brisbane & \makecell[l]{Brisbane \\ Entertainment Centre} \\
        6th February, 2010 & Fearless Tour & Australia & Sydney & Acer Arena \\
        \vdots & \vdots & \vdots & \vdots & \vdots \\
        7th December, 2024 & The Eras Tour & Canada & Vancouver & BC Place \\
        8th December, 2024 & The Eras Tour & Canada & Vancouver & BC Place \\
        \bottomrule
    \end{tabular}
    \vspace{0.5em}
    
    \noindent\textit{(Full table contains 533 entries and is truncated for clarity)}
    \end{tcolorbox}

    \vspace{0.5em}

    \begin{tcolorbox}[
        enhanced,
        title=Evaluation Criteria,
        colback=violet!5!white,
        colframe=violet!75!black,
        fonttitle=\bfseries,
        attach boxed title to top left={yshift=-2mm, xshift=3mm},
        boxed title style={
            colback=violet!75!black,
        }
    ]
    \small
    \textbf{Unique Columns:} \texttt{["Date"]}
    \vspace{0.5em}

    \textbf{Required Columns:} \texttt{["Date", "Concert’s English Name", "Host Country", "Host City", "Host Venue"]}
    \vspace{0.5em}
    
    \textbf{Evaluation Pipeline:}
    \begin{itemize}[leftmargin=*, noitemsep]
        \item \textbf{Date}:
        \begin{itemize}[leftmargin=*, noitemsep]
            \item \textit{Pre-process}: \texttt{["norm\_str"]}
            \item \textit{Metric}: \texttt{["exact\_match"]}
        \end{itemize}
        \item \textbf{Concert’s English Name}:
        \begin{itemize}[leftmargin=*, noitemsep]
            \item \textit{Pre-process}: \texttt{["norm\_str"]}
            \item \textit{Metric}: \texttt{["exact\_match"]}
        \end{itemize}
        \item \textbf{Host Country}:
        \begin{itemize}[leftmargin=*, noitemsep]
            \item \textit{Pre-process}: \texttt{["norm\_str"]}
            \item \textit{Metric}: \texttt{["exact\_match"]}
        \end{itemize}
        \item \textbf{Host City}:
        \begin{itemize}[leftmargin=*, noitemsep]
            \item \textit{Pre-process}: \texttt{["norm\_str"]}
            \item \textit{Metric}: \texttt{["llm\_judge"]}
            \item \textit{Criterion}: It is sufficient if the semantics are approximately the same as the reference answer or if they point to the same entity. There is no need for a word-for-word correspondence.
        \end{itemize}
        \item \textbf{Host Venue}:
        \begin{itemize}[leftmargin=*, noitemsep]
            \item \textit{Pre-process}: \texttt{["norm\_str"]}
            \item \textit{Metric}: \texttt{["llm\_judge"]}
            \item \textit{Criterion}: It is sufficient if the semantics are approximately the same as the reference answer or if they point to the same entity. There is no need for a word-for-word correspondence.
        \end{itemize}
    \end{itemize}
    \end{tcolorbox}
    
    \caption{A visually enhanced example of a task from our benchmark. The task is separated into a styled \textbf{Task Prompt} box, a \textbf{Ground-Truth} box, and an \textbf{Evaluation Criteria} box.}
    \label{fig:example_with_eval}
\end{figure}

\subsection{Benchmark Composition and Statistics} 

\begin{figure}[t]
    \centering
    \includegraphics[width=1.0\textwidth]{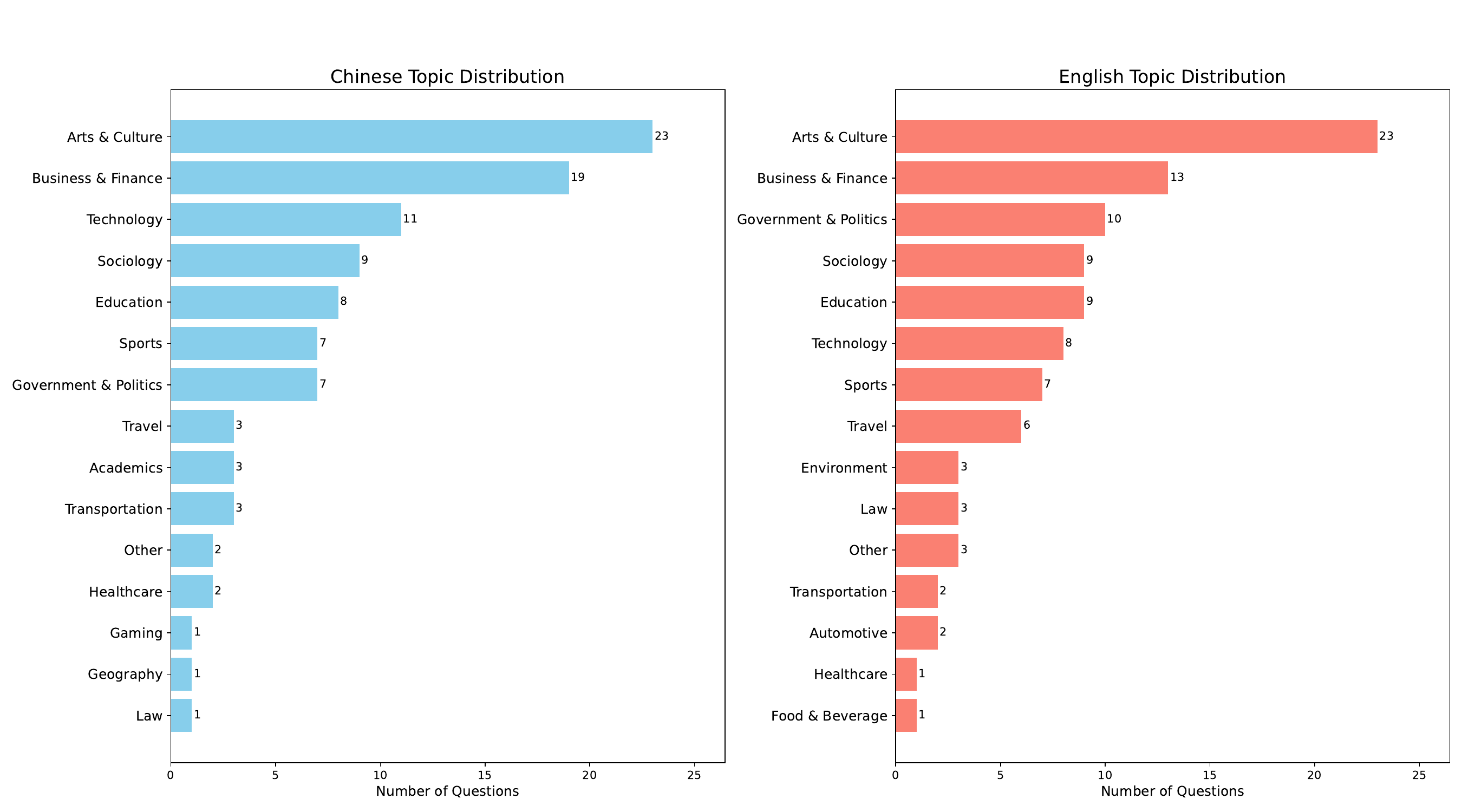}
    \caption{Distribution of the 18 distinct topics across the 200 tasks in the WideSearch benchmark, ensuring broad domain coverage.}
    \label{fig:widesearch_topic_distribution}
\end{figure}

The rigorous curation pipeline culminates in the final WideSearch benchmark, which comprises 200 high-quality tasks. For robust cross-lingual evaluation, these tasks are distributed equally between English and Chinese (100 tasks per language). Furthermore, to ensure broad applicability and mitigate domain-specific biases, the tasks are methodically balanced across 18 diverse topics, as detailed in Figure \ref{fig:widesearch_topic_distribution}.

To quantitatively substantiate the complexity inherent in our benchmark, we conduct a detailed human annotation study with 30 participants. This evaluation is performed on a representative subset of 100 tasks, drawn equally from the Chinese and English pools (50 tasks each). Annotators are given ample time and instructed to complete each task independently to achieve the highest possible accuracy. However, we acknowledge that due to the numerous data points required for each complex task, even a diligent human annotator may commit inadvertent errors in a single session. To mitigate the impact of such potential errors and establish a robust performance ceiling, we implement a dual-annotation protocol. Each task is independently completed by two annotators, and we exclusively utilize the data from the annotator who achieved higher accuracy. This rigorous methodology ensures that our complexity metrics are grounded in high-quality, successful task completions.

\begin{figure}[t]
    \centering

    \begin{subfigure}{0.49\textwidth}
        \centering
        \includegraphics[width=\textwidth]{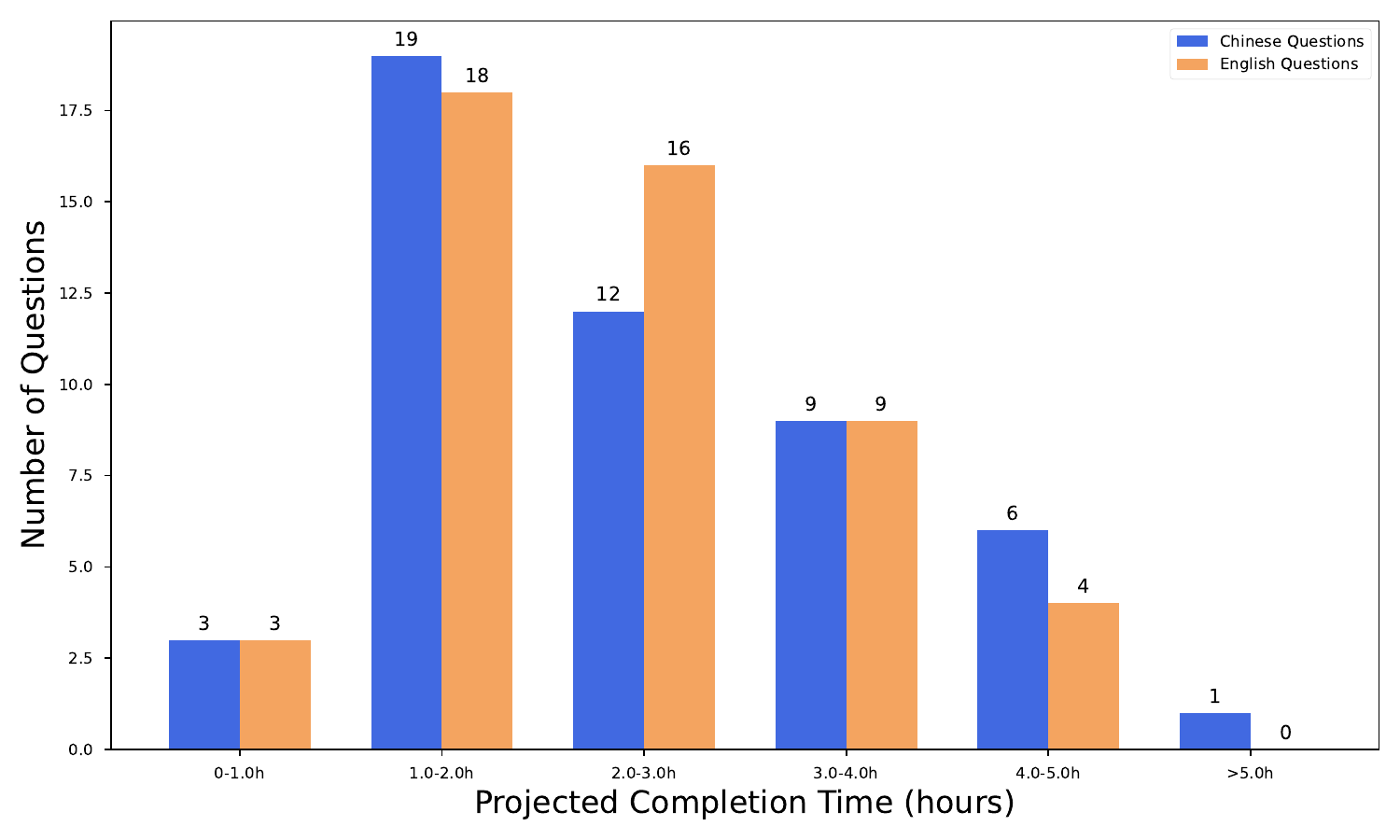}
        \subcaption{Distribution of task completion time.}
        \label{subfig:completion_time}
    \end{subfigure}
    \hfill 
    \begin{subfigure}{0.49\textwidth}
        \centering
        \includegraphics[width=\textwidth]{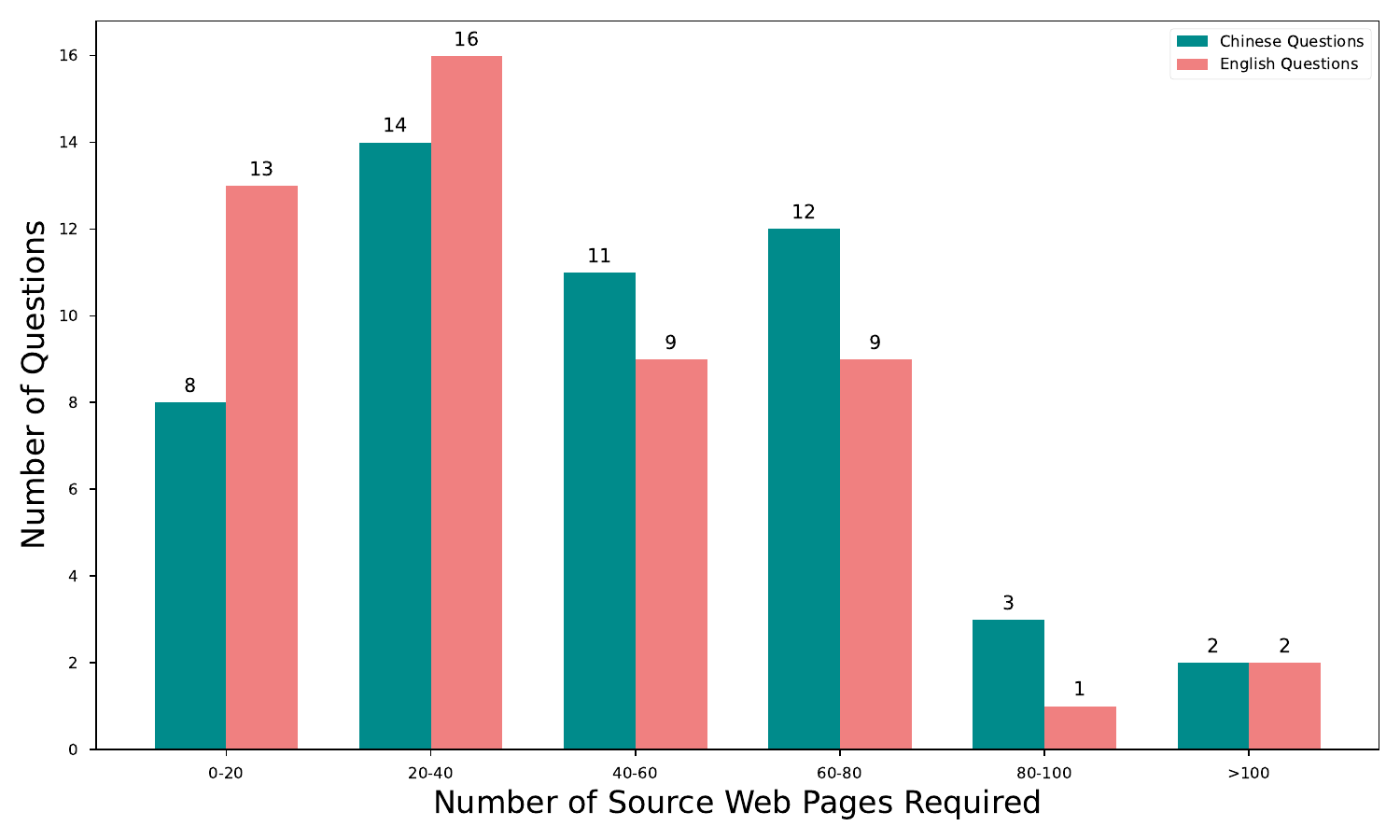}
        \subcaption{Distribution of source web pages consulted.}
        \label{subfig:web_pages}
    \end{subfigure}

    \caption{Statistical distributions of key complexity metrics from our human annotation study. Both charts compare performance on Chinese and English tasks, showing (a) the time required for completion and (b) the breadth of research needed.}
    \label{fig:complexity_distributions}
\end{figure}

Our analysis focuses on several key indicators. The first two metrics, derived from the human study, measure the procedural effort. As illustrated in Figure \ref{subfig:completion_time}, we report the distribution of human completion times. It directly reflects the significant cognitive and temporal investment demanded by each task, with an overall average completion time of 2.33 hours. This is remarkably consistent across languages, with English tasks averaging 2.29 hours and Chinese tasks averaging 2.37 hours. This metric is comprehensive, encapsulating the entire workflow from initial query comprehension, through multi-step searching and information synthesis, to final result validation.

Furthermore, to quantify the procedural depth, Figure \ref{subfig:web_pages} shows the number of unique source web pages that annotators consulted. The breadth of research required is extensive; on average, annotators need to consult 44.10 unique web pages per task (48.74 for Chinese and 39.46 for English). Annotators are not limited in their choice of search tools. Crucially, this number represents not a theoretical minimum, but the actual breadth of research performed, including the cross-verification of facts across multiple sources to ensure accuracy. It therefore serves as a strong proxy for the non-trivial nature of the information-seeking process.

Finally, to characterize the informational scope across the \textit{entire} benchmark, Table \ref{tab:volume_distribution} presents the distribution of answer data volume. This metric reflects the amount of factual information that must be synthesized and structured to provide a complete and correct solution.

\input{tables/volume}

\subsection{Evaluation Framework and Metrics}

To facilitate an accurate, scalable, and nuanced assessment of agent performance, we develop a comprehensive evaluation framework centered around an automated scoring pipeline. This framework is designed to handle the structured nature of our ground-truth data—which is stored in tables—and to address the inherent complexities of natural language responses.

\subsubsection{Automated Evaluation Pipeline}

To ensure a robust, scalable, and fully automated evaluation, we formalize the scoring process as a task of table alignment and cell-wise verification. Our method is executed through a hybrid pipeline that synergistically combines deterministic rule-based checks with semantic judgments from a large language model (LLM-as-a-judge). The entire process is underpinned by meticulously annotated ground-truth data, which includes pre-defined primary keys for row alignment and column-specific evaluation methods for cell-wise scoring. We use the GPT-4.1-2025-04-14 as the default judge LLM.


The evaluation pipeline, as depicted in Figure \ref{fig:curation_pipeline}, consists of the following stages:

\textbf{Data Preparation and Syntax Validation:} For each ground-truth table, we pre-define a primary key—a single column or a composite of multiple columns—to uniquely identify each row. This key is strictly enforced through one-shot examples in the agent's prompt. The evaluation begins with a critical syntax validation. An agent's response is immediately assigned a score of zero if it is not a valid Markdown table that can be correctly parsed, or if its column headers do not match the ground truth in number and name. Please note that the string of the generated column name may have slight differences from the ground truth, but it must be semantically identical. These slight differences are allowed. We use the mapping prompt in the Appendix \ref{app:eval_details} to align these differences.

\textbf{Normalization and Alignment:} Responses that pass the initial check undergo a series of normalization procedures, such as removing extraneous whitespace and standardizing special characters. For the columns corresponding to the primary keys, we also need to use a mapping prompt to align the entities in the response with the entities in the ground truth. Otherwise, we may not be able to execute a join operation on the two tables. The predicted table is then aligned with the ground-truth table by performing a join operation on the pre-defined unique keys. This alignment allows us to identify matched rows, as well as false positives and false negatives.

\textbf{Hybrid Item-level Scoring:} For each pair of aligned rows, we iterate through the corresponding cells. The evaluation method for each cell is dictated by its column's pre-annotated type, enabling nuanced and accurate scoring. Our framework supports a comprehensive set of categories:

\begin{itemize}
    \item \textbf{Exact Match}: For strings where absolute precision is paramount.

    \item \textbf{Numerical Approximation}: To validate numbers while allowing for minor, acceptable floating-point or formatting variations.

    \item \textbf{Date Matching}: To semantically compare dates that may appear in different but equivalent formats (e.g., "July 4th, 2024" vs. "2024-07-04").

    \item \textbf{URL Matching}: To normalize and validate the correctness of web links.

    \item \textbf{LLM-as-a-judge}: Reserved for complex cases with high lexical variation (such as translated names or nuanced descriptions) that require semantic understanding for a fair assessment. The prompt of the LLM-as-a-judge is shown in Appendix \ref{app:eval_details}.
\end{itemize}

\subsubsection{Evaluation Metrics}
\label{sec:eval_metric}
The results from the pipeline are aggregated into a suite of metrics for a multi-faceted analysis:

\begin{itemize}
    \item \textbf{Success Rate (SR)}: Our primary and most stringent metric is the Success Rate. A task is considered completed if and only if the agent-generated Markdown table is a perfect match to the ground-truth table, including all content and structure. While SR provides an unambiguous measure of overall task completion, its binary, all-or-nothing nature is often too coarse for a detailed analysis, especially given the large number of data points in each task.
    \item \textbf{Row-level F1 Score}: To overcome the limitations of SR, we introduce a row-level F1 score. In this scheme, each row of the table is treated as a fundamental unit of information, representing a complete record or entity. We compute the precision, recall, and F1 score by comparing the set of rows in the predicted table \(P_{rows}\) against the set of rows in the ground-truth table \(G_{rows}\). This metric assesses the agent's ability to retrieve and correctly structure complete entries.
    \item \textbf{Item-level F1 Score}: For an even more granular assessment, we employ an item-level F1 score. Here, each cell or data point within the table is considered the basic unit for comparison. We calculate precision, recall, and F1 score based on the multiset of items in the predicted table \(P_{items}\) and the ground-truth table \(G_{items}\). This metric evaluates the agent's fine-grained accuracy in extracting specific pieces of information, making it particularly useful for identifying partial successes or minor errors.
\end{itemize}

Furthermore, to provide a more comprehensive evaluation, we perform N independent runs for each task and report performance using three aggregation strategies:

\begin{itemize}
\item \textbf{Avg@N}: This metric measures the agent's average performance. For each of the N runs per task, we record the binary outcome for Success Rate (1 for success, 0 for failure), as well as the Row-level and Item-level F1 Score. The Avg@N for each metric type is the arithmetic mean of these N values. For SR, this average represents the success rate over N trials for a given task.
\item \textbf{Pass@N}: This metric captures the agent’s peak capability on the Success Rate. For each task, we determine whether the task was solved successfully in at least one of the N runs. The overall Pass@N score is the percentage of tasks that meet this criterion across the dataset.
\item \textbf{Max@N}: For Row-level and Item-level metrics, we report Max@N. For each task, we take the single highest F1 score achieved across the N runs. The overall Max@N score is the average of these maximum values over all tasks in the dataset.
\end{itemize}

%% file: tables/volume.tex
\begin{table}[t]
\centering
\caption{Projected Distribution of Answer Data Volume. Data volume is defined as the number of discrete factual data points (e.g., rows multiplied by columns in a result table) required for a complete answer.}
\label{tab:volume_distribution}
\begin{tabular}{lcc}
\toprule
\textbf{Data Volume Range} & \textbf{Chinese Questions} & \textbf{English Questions} \\
\midrule
0 - 100                    & 0                        & 2                        \\
100 - 1000                 & 59                        & 77                        \\
1000 - 2000                & 17                        & 13                        \\
2000 - 3000                & 9                        & 5                        \\
3000 - 4000                & 2                        & 0                        \\
4000 - 5000                & 7                        & 0                        \\
> 5000                     & 6                        & 3                        \\
\midrule
\textbf{Average Volume}    & \textbf{2001.2}               & \textbf{938.6}               \\
\bottomrule
\end{tabular}
\end{table}

%% file: sections/approach.tex


%% file: sections/experiments.tex
\section{Experiments}

\subsection{Experimental Setup}

To comprehensively evaluate agent capabilities on our WideSearch benchmark, our experimental design targets three distinct aspects: the performance of the single-agent framework, the effectiveness of a multi-agent framework, and a comparative benchmark against leading end-to-end systems. For our modular agent architectures (Single and Multi-Agent), each agent is equipped with a standardized toolset comprising a search tool (Bing Search API) and a webpage reading tool. To test the native agentic capabilities, we use the most naive agent architecture (single-agent and multi-agent), without carefully designing the system prompt or any complex workflows. For example, in the single-agent mode, we do not mandate that the model must self-reflect; in the multi-agent mode, we do not dictate how detailed the task decomposition should be. We provide the details of the agents and tools in Appendix \ref{app:agent_details}, and the API identifiers for all models used are listed in Appendix \ref{sec:appendix_models}.

\textbf{Single Agent.} Our first objective is to measure the capability of the single-agent framework under different LLMs. In this mode, a single LLM, equipped with the aforementioned tools, is responsible for the entire task lifecycle—from planning and information seeking to synthesizing the final answer. This setup serves as a crucial baseline to assess the intrinsic problem-solving abilities of each model. The foundation models evaluated in this configuration are: DeepSeek-R1 \cite{guo2025deepseek}, Doubao-Seed-1.6 (Thinking) \cite{seed2025seed1}, Doubao-Seed-1.6 (Non-Thinking) \cite{seed2025seed1}, Claude Sonnet 4 (Thinking) \cite{Claude4}, Gemini 2.5 Pro \cite{Google_Gemini25Pro}, Kimi K2 \cite{team2025kimi}, and OpenAI o3 \cite{OpenAI_o3}.

\textbf{Multi-Agent Framework.} Recognizing the inherent parallelism in WideSearch tasks, we evaluate a multi-agent framework to test the effectiveness of a "divide-and-conquer" strategy. The framework consists of a main agent that decomposes the query and aggregates the results, and multiple sub-agents that execute the sub-tasks in parallel. To ensure a direct comparison, we test each of the foundation models listed above within this framework, allowing us to systematically measure the performance impact of the architecture itself versus the single-agent paradigm.

\textbf{End-to-End Systems.} Our third objective is to contextualize performance against state-of-the-art commercial solutions. We initially intended to evaluate dedicated "DeepResearch" systems. However, we observed that these systems often struggle to adhere to specific instructions; instead of generating a single, correctly formatted Markdown table, they frequently return a long-form report accompanied by multiple tables. This non-compliance makes automated programmatic evaluation difficult. Consequently, we shift our focus to benchmarking the integrated web-browsing mode of leading commercial systems. For this comparison, we specifically evaluate Gemini 2.5 Pro, Claude Sonnet 4 (Thinking), and OpenAI o3.

\textbf{Human Evaluation.} The process of annotating the ground truth for WideSearch is a very arduous task, requiring multiple annotators to repeatedly search and cross-validate information. To test the ability of a single person to solve WideSearch tasks, we randomly selected 10 questions in Chinese and 10 in English. We then invited an additional 10 annotators to participate in an experiment, with each person working on two questions individually. Each participant was allowed to use any tool (including any existing AI assistants) and take as much time as needed until they were confident that their answer was complete.


\subsection{Main Results}

\input{tables/main_results}

We report the main experiment results in Table \ref{tab:main_results}. The conclusions are obtained as follows:

\textbf{Existing models still lack the advanced agentic abilities.} Current advanced large language models show fundamental weaknesses when performing large-scale information-seeking tasks, with failures stemming from fundamental cognitive deficits beyond simple search inaccuracies. They exhibit poor planning by struggling to break down complex questions into comprehensive sub-queries, which leads to incomplete information seeking. Furthermore, they lack reflection and fail to dynamically adjust their search strategy when initial attempts are unsuccessful, often giving up or answering with insufficient data instead of trying new methods. Even when they successfully find relevant information, they demonstrate faulty evidence use by misinterpreting or misattributing the content. These basic deficiencies in planning, dynamic adjustment, and reasoning are the primary reasons for their extremely low success rates on such complex tasks.

\textbf{Multi-agent mode outperforms the single-agent mode on WideSearch.} The multi-agent framework, using a "divide-and-conquer" strategy, consistently and significantly outperforms the single-agent mode on WideSearch tasks by more effectively addressing their inherent breadth. Although absolute success rates are low for both, the multi-agent system shows a distinct advantage in F1 scores, which measure partial correctness. This superior performance is due to its architecture, where a planner decomposes a broad query into parallel sub-tasks assigned to different agents. This parallel search and division of labor not only improves the breadth and efficiency of information seeking but also mimics the specialized, collaborative process of human expert teams, making the framework better suited for complex, wide-ranging searches.

\textbf{Current commercial AI assistants cannot yet seek information at a large scale.} Although top commercial AI models have some information retrieval capabilities in their integrated web Browse modes, the results of the WideSearch test show that they still struggle with information-seeking tasks that require large-scale and high-precision output. Several leading commercial models tested in the experiment, including Gemini 2.5 Pro, Claude Sonnet 4, and OpenAI o3, hover around a 5\% table-level success rate. Furthermore, in the early stages of the experiment, we found that some specialized "DeepResearch" systems even had difficulty following precise instructions. They tend to generate lengthy reports rather than the single, well-formatted table required by the task. It demonstrates that the design of current mainstream AI assistants has not yet been optimized for large-scale, systematic information integration and verification, and they lack the stability and precision required to become reliable productivity tools.

\textbf{Even humans cannot achieve a high success rate in single-player mode.} Experimental results show that even when given ample time and access to any tools, the success rate for a single individual completing the task independently is merely 20\%. This outcome highlights the inherent difficulty of the task itself. A key characteristic of WideSearch tasks is the extreme density of data points; a complete answer may contain thousands of individual facts. Under these circumstances, any minor error—whether it's an extra, a missing, or an incorrect piece of data—results in the failure of the entire task according to the strict success criteria. The construction of the "ground truth" for the benchmark is an incredibly arduous task, requiring multiple annotators to perform several rounds of repeated searches, revisions, and cross-validations to ensure its accuracy. Hence, requiring a single agent (human or AI) to flawlessly collect, integrate, and verify all information in a single attempt is an exceptionally high bar. This demonstrates the challenging yet reasonable nature of WideSearch as a benchmark for evaluating the robustness and completeness of search agents.

For a detailed domain-level performance analysis, please refer to Appendix~\ref{sec:domain_analysis}.

\subsection{Consistency with Human Evaluation}

To validate the stability and reliability of our proposed automated evaluation pipeline, we conduct a consistency analysis against human assessment. For this analysis, we curate an evaluation set of 200 responses by randomly selecting one output from the pool of commercial agentic systems for each task in WideSearch. These selected responses are then meticulously annotated by human experts to determine their item-level correctness against the ground truth.

Subsequently, we utilize our evaluation pipeline, employing different models as judges, to assess the same set of responses. The primary objective is to measure the degree of agreement between our automated pipeline's judgments and the human-annotated labels.

The results of this comparison are presented in Table \ref{tab:human_consistency}. As shown, the consistency between our pipeline's evaluation and human assessment is exceptionally high, exceeding 97.8\% for all tested judge models, including both thinking and non-thinking variants. This high level of correlation underscores the effectiveness and reliability of our proposed evaluation methodology. Furthermore, it reinforces the objective nature of the WideSearch benchmark, demonstrating that performance can be assessed accurately and consistently without being subject to the variability of human evaluation.

\begin{table}[t]
\caption{Consistency between our evaluation pipeline using different judge models and human evaluation. (\%)}
\centering
\begin{tabular}{l|c}
\hline
\textbf{Judge Model} & \textbf{Consistency with Human Evaluation} \\
\hline
OpenAI o4-mini & 98.3 \\
Gemini 2.5 Pro & 98.1 \\
GPT-4.1 & 98.0 \\
Doubao-Seed-1.6 (Non-Thinking) & 97.8 \\
\hline
\end{tabular}
\label{tab:human_consistency}
\end{table}

\section{Analysis}

\label{sec:analysis}

To gain a deeper understanding of the core challenges that current models face on the WideSearch benchmark, we conduct a systematic analysis of the experimental data and failure cases. We categorize the primary failure modes into two main groups: 1) Challenges in Advanced Agentic Capabilities, which reflect fundamental deficiencies in complex cognitive skills such as planning, reasoning, and synthesis; and 2) Basic Failure Modes, which arise from the model's inability to reliably execute explicit instructions or tool-use protocols. This classification not only highlights the technical bottlenecks of current search agents but also provides clear directions for future research.

\subsection{Challenges in Advanced Agentic Capabilities}

In large-scale information gathering scenarios (i.e., WideSearch), the balance between Precision and Recall remains a core challenge, which is consistent with the challenges faced in traditional information retrieval tasks. Experiments indicate that the model's performance is far from optimal, both at the row-level and item-level evaluations. A particularly prominent phenomenon is that Recall is significantly lower than Precision across all test subsets, as shown in Table \ref{tab:app_experiments}. This finding reveals a critical deficiency in the current model's ability to capture comprehensive information, identifying that "inadequate recall" is the primary bottleneck constraining its performance. Through an in-depth review and manual analysis of the Agent's reasoning process, we have identified the following key failure patterns:

\noindent \textbf{Incomplete Query Decomposition.} When faced with complex, multi-faceted search topics, LLMs often fail to fully decompose the user's intent into a comprehensive and complementary set of sub-queries. This leads them to miss key constraints or scopes of inquiry during multi-turn searches, failing to gather sufficient information to formulate a final answer. This issue exposes a weakness in the model's capacity for complex task planning and structured decomposition. For detailed case studies, please refer to Appendix Figure \ref{fig:incomplete_query}.

\noindent \textbf{Lack of Reflection and Iterative Refinement}: When an initial tool call returns no results or insufficient information, an ideal agent should be able to "reflect" on the cause of failure and proactively adjust its search strategy (e.g., by reformulating keywords or broadening/narrowing search criteria). However, we find that even advanced large reasoning models lack this dynamic adjustment mechanism. They tend to abandon the search after an initial failed attempt and proceed to answer based on incomplete information or their internal knowledge, reflecting a deficiency in critical thinking and adaptive planning. For detailed case studies, please refer to Appendix Figure \ref{fig:reflection}.

\noindent \textbf{Failure in Evidence Utilization.} This failure occurs when the agent does not correctly ground its final answer in the evidence it retrieves, revealing a critical gap between information retrieval and generation. This deficiency typically appears in two ways: the agent either misinterprets or disregards the content of a relevant source, or it fails to validate a source's context and relevance, thereby misapplying factually correct but inappropriate information. Both issues stem from a fundamental breakdown in evidence evaluation. For detailed case studies, please refer to Appendix Figure \ref{fig:evidence}.

\noindent \textbf{Knowledge Hallucination and Factual Inconsistency.} When the search engine fails to return any relevant information, LLMs sometimes just use their internal knowledge. This frequently leads to "hallucinations," where the model fabricates non-existent facts or provides incorrect information that conflicts with established knowledge. This problem underscores the critical importance and challenge of strictly "grounding" the outputs of LLMs in externally verifiable sources. For detailed case studies, please refer to Appendix Figure \ref{fig:hallucination}.

\subsection{Basic Failure Modes}

In addition to the sophisticated agentic deficiencies described above, we also catalog a series of more basic failures where the model failed to generate the desired output. These errors often lead directly to the termination of the task workflow.

\noindent \textbf{Tool Invocation Error.} This is one of the most common failures, typically caused by the model generating incorrect parameter formats, omitting necessary arguments, or attempting to call a non-existent tool, leading to an API call failure.

\noindent \textbf{Output Formatting Error.} A subset of LLMs fails to strictly adhere to the output format requirements specified in the instructions, such as failing to generate a Markdown table or producing a malformed one.

\noindent \textbf{Context Length Exceedance.} The task is prematurely terminated because the model generated overly verbose intermediate steps or became trapped in ineffective loops, causing the total input to exceed the model's maximum context window.

\noindent \textbf{Response Refusal.} The model exhibits refusal behaviors for a subset of queries. We have identified two primary patterns of refusal: 1. The model perceives ambiguity in the user's question and consequently requests further clarification to narrow the scope of the inquiry. 2. The model deems the required information too extensive to be presented in a single output, leading to a direct refusal to respond.

\section{Test-time Scaling}
Allocating more compute resources during testing is a common method for probing the upper limits of a model's performance. In the experiments in this section, we use Kimi K2 as the foundation model, equipped with Search and Web Browse tools. Based on the single-agent mode, we attempt each question N times (where N is expanded from 1 to 128) and record the Success Rate (Pass@N), Row-level F1 Score (Max@N), and Item-level F1 Score (Max@N).

\begin{wrapfigure}{r}{0.5\linewidth}
  \vspace{-10pt} 
  \centering
  \includegraphics[width=\linewidth]{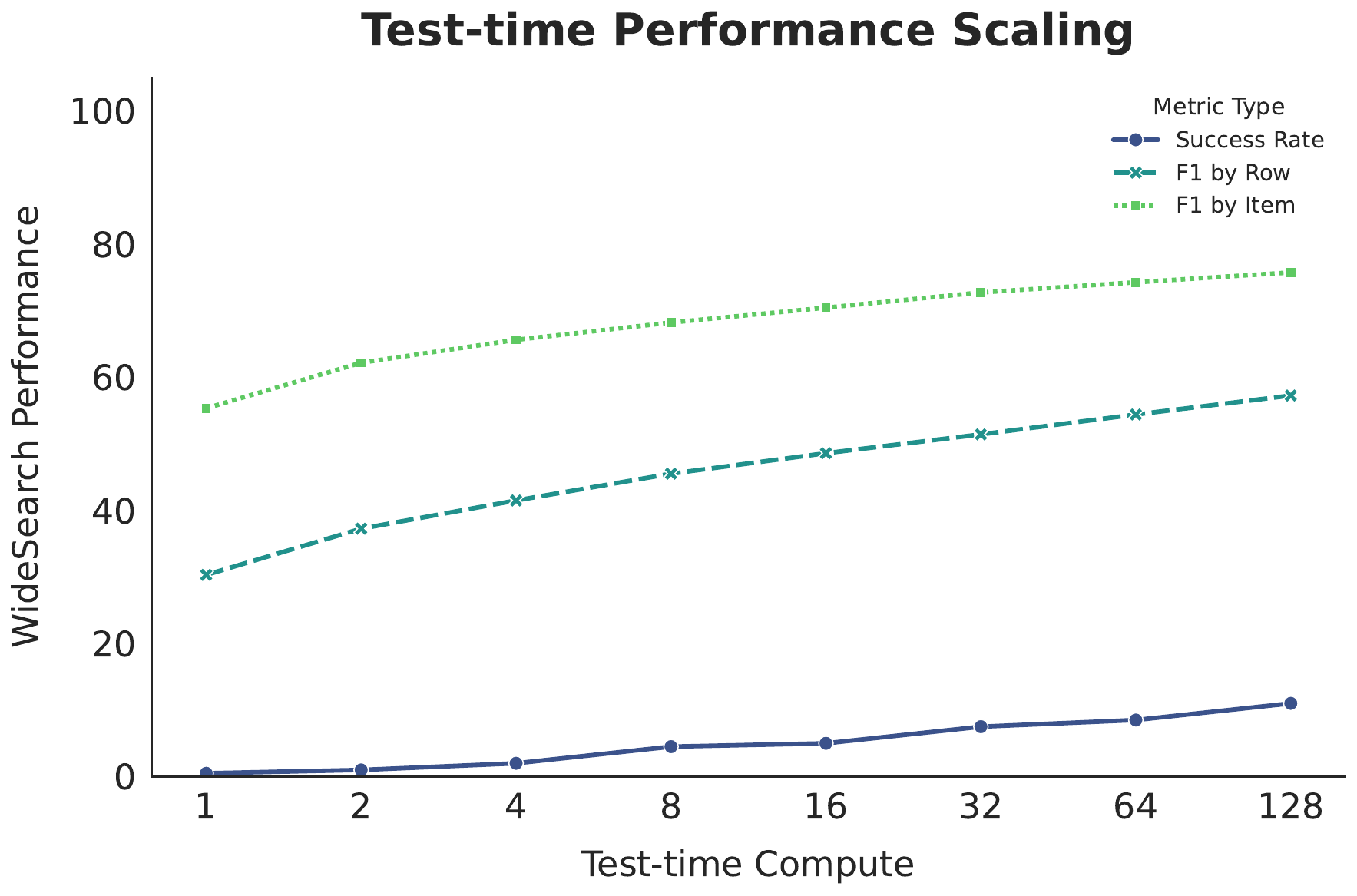}
  \caption{Test-time scaling experiments. We report the Pass@N for Success Rate, Max@N for Row-level, and Item-level F1 score.}
  \label{fig:tts}
  \vspace{-10pt} 
\end{wrapfigure}

As shown in Figure \ref{fig:tts}, the Item-level F1 score shows a significant improvement as the number of attempts increases. With the compute volume of 128 attempts, it even reaches a level close to 80 points. This fully indicates that a single information-seeking action within WideSearch is not a particularly difficult task. Corresponding to real-world human scenarios, finding a piece of basic information is not an exceedingly difficult task for a person, provided enough time is spent. Quite the opposite, even with 128 attempts, the table-level score only reaches a level below 20 points. This strongly suggests that large-scale information retrieval on a fixed topic is an extremely difficult task. It requires not only comprehensive but also accurate search results. For a task with 5,000 atomic pieces of information, even if you find 4,999 correct pieces, the entire task is considered a failure if you retrieve one extra, one fewer, or one incorrect piece of information. Even for humans, completing this task requires multiple annotators to perform several rounds of repeated revisions (which is how the ground truth table for each question was annotated).

Given the characteristics of the WideSearch, we view the optimization of a multi-agent architecture as an important future research direction. Multiple agents can conduct parallel searches and perform mutual cross-validation, a process that aligns highly with the cognitive process of human annotation.

%% file: tables/main_results.tex
\begin{table*}[t]
\centering
\caption{Main results on the WideSearch benchmark. We report Success Rate (SR), Row-level F1, and Item-level F1 for all evaluated systems. All scores are reported as percentages (\%).}
\label{tab:main_results}
\resizebox{1\linewidth}{!}{
\begin{tabular}{l cc cc cc cc}
\toprule
\textbf{Model / System} & \multicolumn{2}{c}{\textbf{Success Rate (\%)}} & \multicolumn{2}{c}{\textbf{Row F1 Score (\%)}} & \multicolumn{2}{c}{\textbf{Item F1 Score (\%)}} & \multicolumn{2}{c}{\textbf{\# Tool Calls}}\\
\cmidrule(lr){2-3} \cmidrule(lr){4-5} \cmidrule(lr){6-7} \cmidrule(lr){8-9}
& Avg@4 & Pass@4 & Avg@4 & Max@4 & Avg@4 & Max@4 & Search & Web Browse\\
\midrule
\multicolumn{9}{l}{\textit{Single Agent}} \\
Claude Sonnet 4 (Thinking) & 2.3 & 5.0 & 31.7 & 41.9 & 57.9 & 66.7 & 8.20 & 3.42 \\
Gemini 2.5 Pro & 1.5 & 5.0 & 30.0 & 41.4 & 51.0 & 63.6 & 7.48 & 1.58 \\
OpenAI o3 & 4.5 & 9.0 & 34.0 & 44.1 & 52.6 & 62.3 & 13.26 & 5.75\\
Kimi K2 & 1.1 & 3.5 & 29.7 & 41.4 & 54.4 & 65.1 & 10.78 & 2.22\\
DeepSeek-R1 & 0.4 & 1.5 & 20.7 & 31.7 & 41.3 & 55.1 & 2.91 & 1.40\\
Doubao-Seed-1.6 (Thinking) & 2.6 & 5.0 & 30.0 & 44.1 & 48.3 & 63.9 & 22.08 & 1.14\\
Doubao-Seed-1.6 (Non-Thinking) & 1.0 & 3.5 & 27.2 & 39.9 & 49.0 & 62.0& 8.01 & 1.82\\
\midrule
\multicolumn{9}{l}{\textit{Multi-Agent Framework}} \\
Claude Sonnet 4 (Thinking) & 3.6 & 6.5 & 38.5 & 52.2 & 62.2 & 73.1 & 27.64 & 11.6\\
Gemini 2.5 Pro & 2.0 & 6.5 & 33.5 & 44.6 & 57.4 & 66.3 & 20.73 & 4.72\\
OpenAI o3 & 5.1 & 9.5 & 37.8 & 50.5 & 57.3 & 68.9 & 26.72 & 16.29\\
Kimi K2 & 3.0 & 6.5 & 36.2 & 49.6 & 61.2 & 70.7 & 28.79 & 8.85\\
DeepSeek-R1 & 0.8 & 3.0 & 22.9 & 36.6 & 44.3 & 60.3 & 11.81 & 7.02 \\
Doubao-Seed-1.6 (Thinking) & 2.5 & 5.5 & 34.0 & 48.9 & 54.6 & 69.7 & 52.34 & 6.44\\
Doubao-Seed-1.6 (Non-Thinking) & 2.1 & 4.5 & 29.7 & 42.7 & 52.8 & 65.1 & 14.83 & 5.18\\
\midrule
\multicolumn{9}{l}{\textit{End-to-End Systems}} \\
Claude Sonnet 4 (Thinking) & 2.5 & 5.0 & 24.1 & 33.5 & 48.4 & 58.5 & - & - \\
Gemini 2.5 Pro & 4.3 & 8.0 & 36.6 & 45.4 & 59.1 & 67.2 & - & - \\
OpenAI o3 & 3.0 & 5.5 & 23.9 & 36.0 & 45.5 & 56.5 & - & - \\
\midrule
Human & \multicolumn{2}{c}{20.0} & \multicolumn{2}{c}{69.2} & \multicolumn{2}{c}{82.4} & \multicolumn{2}{c}{-} \\
\bottomrule
\end{tabular}
}
\end{table*}

%% file: sections/conclusion.tex
\section{Conclusion}
This research introduces a new benchmark called WideSearch, designed to evaluate the capabilities of LLM-Agent in tasks that require gathering and integrating extensive structured information from the web, a process termed "wide information seeking".  By testing over 10 leading search agent systems—including single-agent, multi-agent frameworks, and end-to-end commercial systems—on the WideSearch benchmark, the research reveals significant shortcomings in current models. The results show that even the most advanced systems have extremely low success rates on table-level tasks, with the top performer achieving only 5\%, while most systems score near 0\%. In-depth analysis reveals that the root cause of failure is not the inability to find individual pieces of information (item-level F1 scores can be high), but rather the difficulty in finding all the atomic information accurately and comprehensively. The core deficiencies in current agent systems lie in a lack of advanced agentic capabilities, such as the inability to decompose complex problems into comprehensive sub-queries, a lack of reflection and iteration after initial search failures, and the failure to correctly utilize retrieved evidence. In summary, the study demonstrates that current search agents have critical flaws in performing large-scale, high-fidelity information gathering tasks. The findings point to future development directions, indicating the need for more sophisticated agent architectures. In particular, multi-agent systems that can simulate human collaboration, such as parallel search and cross-validation, are identified as a promising approach to tackling these complex tasks.

%% file: sections/acknowledgement.tex
\section*{Acknowledgments}

We gratefully acknowledge the contributions of the following individuals to the benchmark's data creation, review, and annotation (sorted alphabetically by first name): 
Caixia Luo, Chengxing Shuai, Chu Yu, Cui Meng, Fang Zhou, Fei Tang, Fen Xu, Hanjie Wu, Hongqiong Tong, Jiahui Huang, Jiaxin Lei, Jiaxin Lü, Jiao Li, Jie Liao, Junjie Huang, Kai Zhang, Kang Fu, Lanshiyu Chen, Lidan Huang, Ling Tang, Lingying Chen, Ping Zeng, Ruiyi Yang, Shasha Wang, Tongshu Yang, Wengen Xiang, Wenjing Zhang, Xiao Chen, Xiangyu Huang, Xin Fang, Xueling Zhao, Yanan Liu, Yangyang Li, Yihan Cheng, Yihui Jiang, Ying Guo, Yongshuai Hao, Yu Peng, Yuan Zhang, Yuwei Zhang, Yue Wang, Zhe Cao, Zhengrong Xie, Zhiyao Duan, Zhijie Liu, Zhijun Jin, and Zhangling Peng.

%% file: sections/appendix.tex
\section{Models and API Identifiers}
\label{sec:appendix_models}

The models used in this benchmark and their corresponding API identifiers (which often include version or date information) are listed in Table \ref{tab:model_api_names}. The "Benchmark Alias" is the shorthand name used to refer to the model within our paper.

It is worth noting that \texttt{Doubao-Seed-1.6 (Thinking)} and \texttt{Doubao-Seed-1.6 (Non-Thinking)} share the same API identifier but are configured with different generation parameters (with the "thinking" feature enabled and disabled, respectively). Similarly, \texttt{Claude Sonnet 4 (Thinking)} represent specific configurations with the "thinking" feature enabled.

\begin{table}[h]
\centering
\caption{Correspondence between Benchmark Aliases and API Identifiers}
\label{tab:model_api_names}
\begin{tabular}{ll}
\hline
\textbf{Benchmark Alias} & \textbf{API Identifier} \\
\hline
\texttt{Kimi K2}                         & \texttt{kimi-k2-250711}                   \\
\texttt{Doubao-Seed-1.6 (Thinking)}                 & \texttt{doubao-seed-1-6-250615}           \\
\texttt{Doubao-Seed-1.6 (Non-Thinking)}    & \texttt{doubao-seed-1-6-250615}           \\
\texttt{DeepSeek-R1}                & \texttt{deepseek-r1-0528}                      \\
\texttt{Claude Sonnet 4 (Thinking)}    & \texttt{claude-sonnet-4-20250514}                   \\
\texttt{OpenAI o3}                  & \texttt{o3-2025-04-16}                    \\
\texttt{Gemini 2.5 Pro}             & \texttt{gemini-2.5-pro-preview-06-05}     \\
\texttt{GPT-4.1}      & \texttt{gpt-4.1-2025-04-14}               \\
\texttt{OpenAI o4-mini}      & \texttt{o4-mini-2025-04-16}               \\
\hline
\end{tabular}
\end{table}

\section{Detailed Experiments}
We present detailed experimental data in Table \ref{tab:app_experiments}, including Success Rate, Row-level Precision, Row-level Recall, Row-level F1, Item-level Precision, Item-level Recall, and Item-level F1 on different subsets of WideSearch.

\input{tables/appendix_experiments}

\section{Agent Framework Details}
\label{app:agent_details}
We provide the system prompt of the Single-Agent framework, the Multi-Agent framework, and the tools prompt for the agents as follows.  The "Create Sub-Agent" is only provided for the Multi-Agent framework.

\tcbset{colback=seedblue!10!white, colframe=seedblue, width=\linewidth, arc=5mm}
\begin{tcolorbox}
\subsection*{Single Agent Prompt}
\# Role

You are an expert in online search. Your task is gathering relevant information using advanced online search tools based on the user's query, and providing accurate answers according to the search results.

\# Task Description

Upon receiving the user's query, you must thoroughly analyze and understand the user's requirements. In order to effectively address the user's query, you should make the best use of the provided tools to acquire comprehensive and reliable information and data. Below are the principles you should adhere to while performing this task:

- Fully understand the user's needs: Analyze the user's query, if necessary, break it down into smaller components to ensure a clear understanding of the user's primary intent.

- Flexibly use tools: After fully comprehending the user's needs, employ the provided tools to retrieve the necessary information.If the information retrieved previously is deemed incomplete or inaccurate and insufficient to answer the user's query, reassess what additional information is required and invoke the tool again until all necessary data is obtained.

\end{tcolorbox}

\tcbset{colback=seedblue!10!white, colframe=seedblue, width=\linewidth, arc=5mm}
\begin{tcolorbox}
\subsection*{Multi Agent Prompt}
\# Role

You are a professional and meticulous expert in information collection and collation. You can fully understand users' needs, skillfully use search tools, and complete the tasks assigned by users with the highest efficiency.

\# Task Description

After receiving users' questions, you need to fully understand their needs and think about and plan how to complete the tasks assigned by users efficiently and quickly.
To help you complete tasks better and faster, I have provided you with three tools:

1. Search tool: You can use the search engine to retrieve information;

2. Link reading tool: a link reading tool that can open links (which can be web pages, PDFs, etc.) and summarize all relevant information on the page according to the requirement description.

3. Sub Agent: The Sub Agent can complete various types of tasks according to the prompt you input. The Sub Agent itself can also use the search tool and the link reading tool. You can split your tasks into multiple sub-tasks according to your own needs, and then create one or more Agents to help you complete these sub-tasks in parallel.

\end{tcolorbox}

\tcbset{colback=seedblue!10!white, colframe=seedblue, width=\linewidth, arc=5mm}
\begin{tcolorbox}
\subsection*{Tool Description}

\subsubsection*{Search Tool}

Description: 

This is a search tool. Enter search queries, and it will return a list of web pages along with their corresponding summary information. Search queries should be concise and clear; complex questions should be broken down into multiple steps and searched step by step. If no useful pages are found, you can adjust the question description (such as reducing qualifiers or changing the search approach) and search again. The quality of search results is related to the language: for Chinese resources, you can try entering Chinese queries; for non-Chinese resources, you can try using English or the corresponding language.

Parameters:
query, count, summary\_type, use\_english

\vspace{5pt}
\subsubsection*{Text Browser View}

Description: 

This is a link reading tool that can open links (which can be web pages, PDFs, etc.) and summarize all relevant information on the page according to the requirement description. This tool can be called to obtain information for all valuable links. Valuable links include but are not limited to the following types: 1. URLs explicitly provided in the task; 2. URLs with relevant summaries provided in search results; 3. URLs contained in the content returned by previous calls to TextBrowserView that are judged to potentially contain useful information. Please try to avoid constructing links out of thin air by yourself.

Parameters:
url, description

\vspace{5pt}
\subsubsection*{Create Sub-Agent}

Description: 

Creates sub-agents that can perform specific tasks based on the input prompt.

Parameters:
sub\_agents: [(prompt$_0$, index$_{0}$), ..., (prompt$_{N}$, index$_{N}$)]

\end{tcolorbox}

\section{Evaluation Details}
\label{app:eval_details}
Sometimes, the column names in the markdown table generated by the model are completely semantically consistent with those in the ground truth, but the surface strings are not entirely the same. We believe that such cases should be considered as correct predictions by the model, so we use the following mapping prompt to map column names with the same semantics to the same string. Similarly, the inner join on the primary key between the two tables also relies on string matching. We also use the mapping prompt to map entities with the same semantics under the corresponding primary keys in the two tables to the same string. We present an example of the mapping below.

\tcbset{colback=seedblue!10!white, colframe=seedblue, width=\linewidth, arc=5mm}
\begin{tcolorbox}
\subsection*{Mapping Prompt}
Your task is to align two vocabularies. The inputs are the vocabulary to be aligned and the reference vocabulary, respectively. Note that you need to perform semantic alignment (not positional alignment). If two strings are the same, they must correspond to each other. These two strings are supposed to represent the same entity, with differences only in the expression forms and formats.

The vocabulary to be aligned is as follows:

\{response\}

The reference vocabulary is as follows:

\{reference\}

The alignment rules are as follows:

List the values in the vocabulary to be aligned one by one. If there is a value in the reference vocabulary that has the same meaning as this value, `transform` should be represented as the value from the reference vocabulary; otherwise, `transform` should be represented as the original value from the vocabulary to be aligned.

Note that `origin` must be taken from the vocabulary to be aligned, keeping the original format, and `transform` must be taken from the reference vocabulary. For example: Some words in the vocabulary to be aligned might be the words in the reference vocabulary with Markdown formatting added, keep the to be aligned format in `origin` and the reference format in `transform`.

For the `origin`, first find the `transform` that is the closest in meaning, and then judge whether they correspond to each other. Those entities not correspond to each other cannot output.

Please output the alignment results in the following format:

```json

\{

    "origin\_str1": "transform\_str1",
    
    "origin\_str2": "transform\_str2"
    
\}

```

\end{tcolorbox}

\tcbset{colback=seedblue!10!white, colframe=seedblue, width=\linewidth, arc=5mm}
\begin{tcolorbox}
\subsection*{Mapping Output Example}
```json

{

    "gemini-2.5-pro (thinking)": "Gemini 2.5 Pro",
    
    "gemini-2.5-flash (thinking)": "Gemini 2.5 Flash-thinking",
    
    "gemini-2.5-flash (non-thinking)": "Gemini 2.5 Flash-non thinking",
    
    "gemini-2.5-flash-lite (thinking)": "Gemini 2.5 Flash-Lite-thinking",
    
    "gemini-2.5-flash-lite (non-thinking)": "Gemini 2.5 Flash-Lite-non thinking",
    
    "claude-3-7-sonnet": "Claude 3.7 sonnet",
    
    "claude-opus-4": "Claude Opus 4",
    
    "claude-sonnet-4": "Claude Sonnet 4",
    
    "o3 (high)": "O3",
    
    "o3-mini (high)": "O3 mini",
    
    "o4-mini (high)": "o4-mini",
    
    "doubao-1.5-pro-thinking": "Seed-Thinking v1.5",
    
    "doubao-1.6": "Seed-Thinking v1.6",
    
    "deepseek-v3": "DeepSeek-V3",
    
    "deepseek-r1-0528": "DeepSeek-R1-0528"
    
}

```

\end{tcolorbox}

For non-primary key columns, some columns cannot be evaluated using rule-based methods and must be evaluated using the llm-as-judge approach. During the data annotation process, we annotated the LLM-as-judge criteria for these columns. To improve the efficiency of the judge, we perform evaluations column by column. The specific llm-as-judge prompt and evaluation examples are as follows:

\tcbset{colback=seedblue!10!white, colframe=seedblue, width=\linewidth, arc=5mm}
\begin{tcolorbox}
\subsection*{LLM-as-Judge Prompt}

You are an expert in grading answers. Your task is to score the responses to a certain question. Below, you will be provided with a set of standard answers, a set of responses to be graded, and specific grading criteria.

Each answer and each response has an index. Please score each pair of answers and responses in this set according to the following methods:

1. The scoring range is from 0 to 1. A score of 1 indicates a completely correct answer. For deduction items, please refer to the specific grading criteria section.

2. After reading the standard answers, responses to be graded, and grading criteria, please first analyze and judge them item by item according to the grading criteria.

3. The score can only be an integer of 0 or 1.

4. After the analysis and judgment, please provide the final scoring results. Each pair should have a score. Output in Markdown JSON format, as shown below:

```json

\{

    "idx\_xxx": score,
    
    "idx\_yyy": score,
    
    ...
    
\}

```

====== criterion-start ======

\{criterion\}

====== criterion-end ======

====== response-start ======

\{response\}

====== response-end ======

Now start scoring. Please make sure to analyze each item step by step before providing the final scoring results.
\end{tcolorbox}

\tcbset{colback=seedblue!10!white, colframe=seedblue, width=\linewidth, arc=5mm}
\begin{tcolorbox}
\subsection*{LLM-as-Judge Output Example}

Let's analyze each answer and response according to the criterion: semantic equivalence is sufficient; answers do not have to correspond word-for-word.

- idx\_0: response = '1mtokens', target = '1m'

    - ""1mtokens"" clearly refers to ""1 million tokens"" which semantically matches ""1m"" (1 million).
    
    - Score: 1

- idx\_1: response = '1mtokens', target = '1m'

    - Same as above.
    
    - Score: 1

- idx\_2: response = '1mtokens', target = '1m'

    - Same as above.
    
    - Score: 1

- idx\_3: response = '128ktokens', target = '200k'

    - ""128ktokens"" (128,000 tokens) does not semantically match ""200k"" (200,000 tokens).
    
    - Score: 0

- idx\_4: response = '200ktokens', target = '200k'

    - ""200ktokens"" corresponds directly to ""200k"".
    
    - Score: 1

- idx\_5: response = '128ktokens', target = '200k'

    - ""128ktokens"" does not correspond to ""200k"".
    
    - Score: 0

- idx\_6: response = 'nan', target = '200k'

    - ""nan"" means there is no answer, which does not match ""200k"".
    
    - Score: 0

- idx\_7: response = '32ktokens', target = '32k'

    - ""32ktokens"" refers to ""32,000 tokens"" and matches ""32k"".
    
    - Score: 1

- idx\_8: response = 'nan', target = '256k'

    - ""nan"" is no answer, does not match ""256k"".
    
    - Score: 0

- idx\_9: response = '128ktokens', target = '64k'

    - ""128ktokens"" does not correspond to ""64k"".
    
    - Score: 0

- idx\_10: response = '128ktokens', target = '64k'

    - Same as above.
    
    - Score: 0

- idx\_11: response = '128ktokens', target = '64k'

    - Same as above.
    
    - Score: 0

Here is the final scoring according to the criteria:

```json

\{

    ""idx\_0"": 1,
    ""idx\_1"": 1,
    ""idx\_2"": 1,
    ""idx\_3"": 0,
    ""idx\_4"": 1,
    ""idx\_5"": 0,
    ""idx\_6"": 0,
    ""idx\_7"": 1,
    ""idx\_8"": 0,
    ""idx\_9"": 0,
    ""idx\_10"": 0,
    ""idx\_11"": 0
    
\}

```
\end{tcolorbox}

\section{Error Analysis and Examples}
\label{sec:appendix_error_analysis}
\counterwithin{figure}{section}
\setcounter{figure}{0}

This section provides detailed examples for each of the failure modes identified in Section \ref{sec:analysis}. We use the trajectories of Gemini-2.5-pro in single-agent mode as examples. Each case includes the user's task, the agent's actions (e.g., search queries and retrieved evidence), and an analysis of the specific error.

\begin{figure}[t]
\centering
\includegraphics[width=0.8\linewidth]{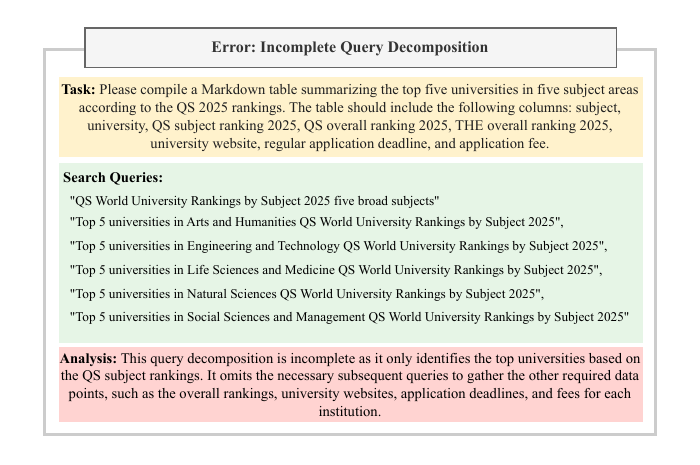}
\caption{An illustration of \textbf{Incomplete Query Decomposition} error. The agent identifies the universities but fails to generate subsequent queries for other required details like application deadlines and fees. Please note that the task description in the figure has been simplified for visualization purposes.}
\label{fig:incomplete_query}
\end{figure}

\noindent \textbf{Case 1: Incomplete Query Decomposition} \
Figure \ref{fig:incomplete_query} illustrates a case of incomplete query decomposition. The agent was tasked with compiling a comprehensive table on the top five universities across five subject areas from the QS 2025 rankings. The required information included not only the rankings but also logistical details such as university websites, application deadlines, and fees. The agent successfully decomposed the initial part of the task, generating specific queries to identify the top universities in each subject area (e.g., "Top 5 universities in Arts and Humanities QS World University Rankings by Subject 2025"). However, it failed to generate the necessary subsequent queries to gather the additional required data points for each university. This omission demonstrates a failure in comprehensive task planning, as the agent did not create a complete set of subtasks needed to fulfill all aspects of the user's request, resulting in a final output with significant information gaps.

\begin{figure}[t]
\centering
\includegraphics[width=0.8\linewidth]{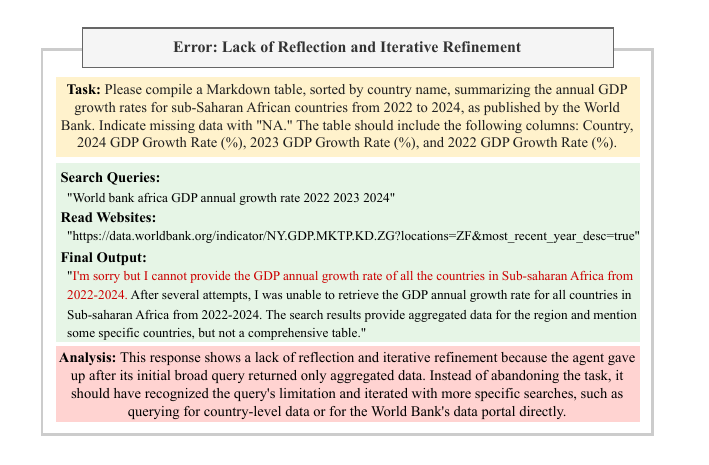}
\caption{An example of the \textbf{Lack of Reflection and Iterative Refinement} error. After an initial broad query returns aggregated data, the agent gives up instead of refining its search to find country-specific information. Please note that the task description in the figure has been simplified for visualization purposes.}
\label{fig:reflection}
\end{figure}

\noindent \textbf{Case 2: Lack of Reflection and Iterative Refinement} \
Figure \ref{fig:reflection} provides an example of the agent's failure to reflect and refine its strategy. The task was to find the annual GDP growth rates for all sub-Saharan African countries from 2022 to 2024, using data from the World Bank. The agent initiated the search with a very broad query: "World bank africa GDP annual growth rate 2022 2023 2024". The search results provided only aggregated data for the entire region, not the country-specific data required. Instead of recognizing this limitation and adapting its approach—for instance, by iterating through individual countries or searching for the specific World Bank data portal—the agent abandoned the task. It concluded that it was unable to retrieve the information, demonstrating a lack of a critical reflective mechanism to diagnose search failures and dynamically adjust its plan.

\begin{figure}[t]
\centering
\includegraphics[width=0.8\linewidth]{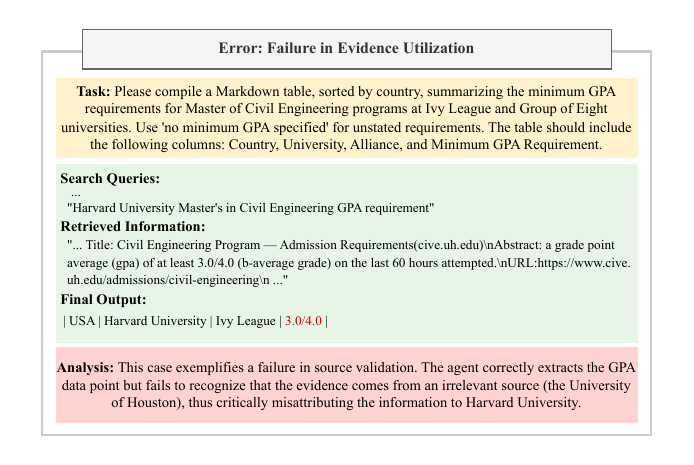}
\caption{An illustration of the \textbf{Failure in Evidence Utilization} error. The agent correctly extracts a GPA requirement but fails to validate its source, misattributing information from the University of Houston to Harvard University. Please note that the task description in the figure has been simplified for visualization purposes.}
\label{fig:evidence}
\end{figure}

\noindent \textbf{Case 3: Failure in Evidence Utilization} \
Figure \ref{fig:evidence} illustrates a critical failure in evidence utilization, specifically in the area of source validation. The agent was asked to find the minimum GPA requirement for a Master of Civil Engineering at Harvard University. It issued a correct query and retrieved a snippet of text stating a GPA requirement of "3.0/4.0". However, the source of this information was cive.uh.edu, the website for the University of Houston, not Harvard. The agent failed to validate the context and relevance of the retrieved evidence, incorrectly attributing the GPA requirement from the University of Houston to Harvard University in its final table. This error highlights a fundamental gap between information retrieval and faithful generation, where the agent does not properly ground its answer in correctly attributed evidence.

\begin{figure}[t]
\centering
\includegraphics[width=0.8\linewidth]{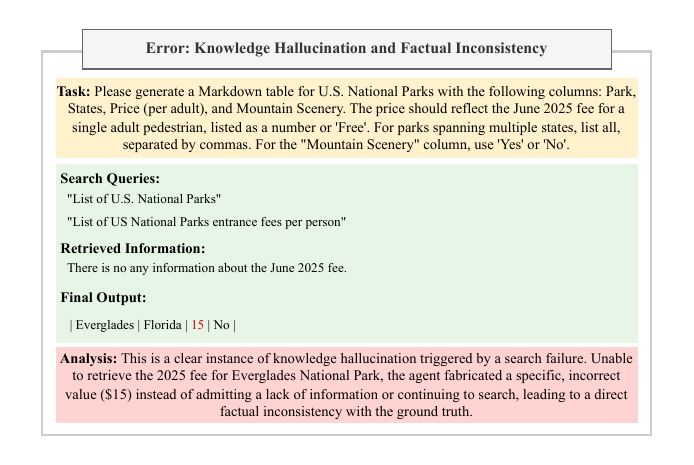}
\caption{An example of \textbf{Knowledge Hallucination and Factual Inconsistency} error. When a search for a future entrance fee returns no results, the agent invents an incorrect value (\$15) instead of stating the information is unavailable. Please note that the task description in the figure has been simplified for visualization purposes.}
\label{fig:hallucination}
\end{figure}

\noindent \textbf{Case 4: Knowledge Hallucination and Factual Inconsistency} \
Figure \ref{fig:hallucination} presents a clear case of knowledge hallucination. The task was to find the entrance fee for U.S. National Parks for June 2025. The agent's search for this future-dated information correctly returned no results. However, when generating the final table for Everglades National Park, instead of indicating that the information was unavailable (e.g., with "NA"), the agent fabricated a specific fee of "\$15". This response was not based on any retrieved evidence but was instead generated from the model's internal parameters, resulting in a factual inconsistency. This case underscores the challenge of ensuring that LLMs remain strictly grounded in external sources, especially when those sources do not contain the requested information.

\section{Domain-Specific Performance Analysis}
\label{sec:domain_analysis}

To move beyond aggregate performance metrics, we conduct a granular analysis to understand how different models and frameworks perform across a variety of specific domains. This approach allows us to identify the strengths and weaknesses of each model and to assess the impact of the multi-agent framework on tasks requiring topic-specific knowledge. The evaluation is performed on both English and Chinese datasets to capture any language-dependent variations in performance.

\begin{figure}[t]
    \centering
    \includegraphics[width=1.0\linewidth]{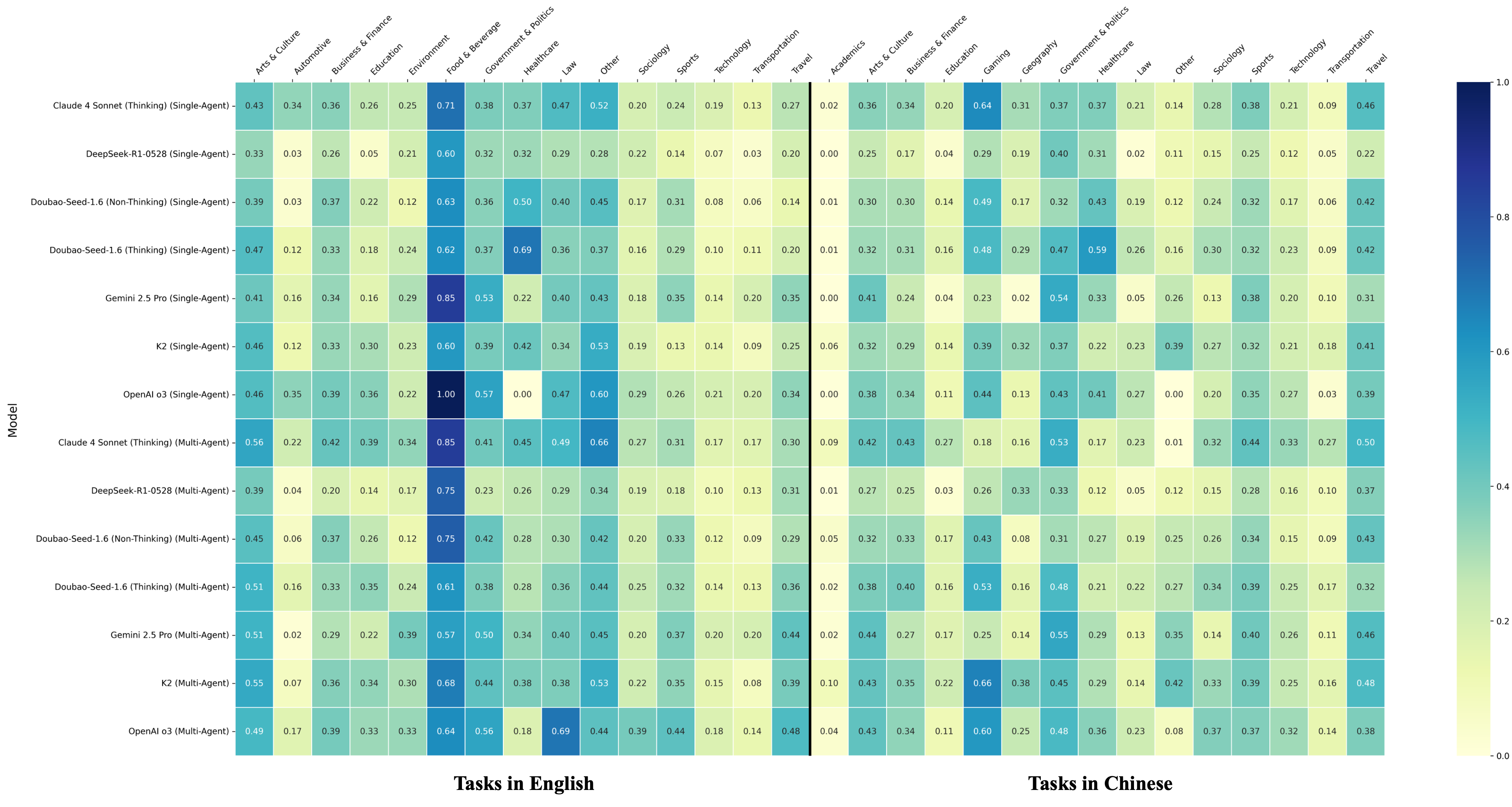}
    \caption{Model performance (Row-level F1 Score) across domains and languages. The heatmap compares F1 scores for various models under single-agent and multi-agent frameworks. Models are evaluated on English and Chinese datasets across multiple domains using an automated pipeline. Darker blue indicates a higher score.}
    \label{fig:widesearch_topic_result}
\end{figure}

Figure~\ref{fig:widesearch_topic_result} presents a detailed heatmap of the row-level F1 scores from this analysis. The figure is structured to facilitate two key comparisons: the performance of different models against each other, and the effectiveness of the single-agent framework (top half) versus the multi-agent framework (bottom half). The domains are grouped by language, with English results on the left and Chinese on the right, separated by a distinct vertical line. As the distribution of questions per domain is uneven, the heatmap serves primarily to identify general patterns rather than to provide a precise statistical comparison.

From the figure, we can draw several key observations. First, the multi-agent framework provides a consistent performance improvement across nearly all models and domains. This suggests that the collaborative approach is effective at enhancing the quality and relevance of the search results, regardless of the topic. Second, certain domains, such as "Academics" and "Transportation," prove to be more challenging for all models, likely due to the need for highly specialized and nuanced information. Finally, the results reveal model-specific aptitudes; for instance, some models show a clear advantage in domains like "Healthcare," while others excel in topics like "Automotive" or "Law." This domain-level insight is crucial for selecting the optimal model and framework for specific real-world applications.

%% file: tables/appendix_experiments.tex
\begin{table*}[t]
\centering
\caption{Detailed experiments results on the WideSearch benchmark.}
\label{tab:app_experiments}
\resizebox{1\linewidth}{!}{
\begin{tabular}{l cc cc cc cc cc cc cc}
\toprule
\textbf{Model / System} & \multicolumn{2}{c}{\textbf{Success Rate}} & \multicolumn{2}{c}{\textbf{Row Precision}} & \multicolumn{2}{c}{\textbf{Row Recall}} & \multicolumn{2}{c}{\textbf{Row F1}} & \multicolumn{2}{c}{\textbf{Item Precision}} & \multicolumn{2}{c}{\textbf{Item Recall}} & \multicolumn{2}{c}{\textbf{Item F1}}\\
\cmidrule(lr){2-3} \cmidrule(lr){4-5} \cmidrule(lr){6-7} \cmidrule(lr){8-9} \cmidrule(lr){10-11} \cmidrule(lr){12-13} \cmidrule(lr){14-15}
& Avg@4 & Pass@4 & Avg@4 & Max@4 & Avg@4 & Max@4 & Avg@4 & Max@4 & Avg@4 & Max@4 & Avg@4 & Max@4 & Avg@4 & Max@4\\
\midrule
\multicolumn{15}{l}{\textit{Single Agent on WideSearch-zh}} \\
Claude Sonnet 4 (Thinking) & 0.25 & 1.00 & 35.91 & 48.63 & 29.08 & 38.11 & 30.19 & 39.73 & 65.05 & 75.99 & 50.84 & 60.51 & 53.76 & 63.19 \\
Gemini 2.5 Pro & 1.00 & 3.00 & 33.98 & 48.04 & 25.68 & 35.13 & 26.95 & 36.96 & 60.49 & 73.46 & 42.35 & 54.45 & 45.57 & 57.26 \\
OpenAI o3-high & 2.00 & 5.00 & 37.10 & 50.97 & 27.02 & 36.23 & 29.30 & 39.31 & 57.31 & 71.48 & 41.45 & 50.19 & 45.19 & 54.46 \\
K2 & 0.25 & 1.00 & 35.53 & 49.23 & 25.89 & 36.68 & 27.79 & 39.03 & 63.15 & 75.05 & 45.05 & 56.07 & 48.81 & 59.64 \\
DeepSeek-R1-0528 & 0.25 & 1.00 & 28.12 & 44.39 & 16.33 & 25.40 & 18.44 & 28.35 & 54.26 & 71.91 & 29.55 & 43.35 & 33.95 & 47.83 \\
Doubao-1.6 & 1.75 & 4.00 & 37.20 & 54.25 & 27.13 & 39.62 & 29.25 & 42.08 & 58.29 & 77.74 & 40.00 & 54.87 & 43.72 & 58.84 \\
Doubao-1.6-non-thinking & 0.50 & 2.00 & 34.48 & 52.17 & 23.67 & 35.00 & 25.56 & 37.41 & 59.97 & 76.31 & 39.14 & 52.01 & 42.87 & 55.79 \\
\midrule
\multicolumn{15}{l}{\textit{Single Agent on WideSearch-en}} \\
Claude Sonnet 4 (Thinking) & 4.25 & 9.00 & 37.01 & 50.55 & 31.97 & 42.69 & 33.18 & 44.08 & 69.83 & 79.13 & 59.67 & 68.32 & 62.02 & 70.27 \\
Gemini 2.5 Pro & 2.00 & 7.00 & 37.66 & 52.62 & 31.37 & 43.74 & 33.05 & 45.82 & 65.57 & 81.19 & 53.67 & 67.38 & 56.38 & 69.97 \\
OpenAI o3-high & 7.00 & 13.00 & 43.04 & 55.30 & 37.48 & 47.26 & 38.70 & 48.84 & 68.28 & 79.30 & 58.07 & 68.31 & 60.03 & 70.08 \\
K2 & 2.00 & 6.00 & 34.71 & 48.42 & 30.36 & 42.34 & 31.54 & 43.68 & 67.08 & 79.51 & 57.63 & 68.42 & 59.91 & 70.52 \\
DeepSeek-R1-0528 & 0.50 & 2.00 & 28.05 & 44.39 & 21.10 & 32.53 & 22.88 & 35.03 & 60.35 & 76.06 & 44.89 & 59.16 & 48.58 & 62.36 \\
Doubao-1.6 & 3.50 & 6.00 & 35.75 & 54.23 & 29.02 & 43.99 & 30.56 & 46.16 & 63.25 & 81.62 & 49.71 & 65.68 & 52.82 & 68.88 \\
Doubao-1.6-non-thinking & 1.50 & 5.00 & 35.84 & 51.98 & 26.90 & 39.88 & 28.86 & 42.31 & 69.29 & 82.92 & 51.23 & 64.86 & 55.06 & 68.17 \\
\midrule
\multicolumn{15}{l}{\textit{Single Agent on WideSearch-all}} \\
Claude Sonnet 4 (Thinking) & 2.25 & 5.00 & 36.46 & 49.59 & 30.52 & 40.40 & 31.69 & 41.90 & 67.44 & 77.56 & 55.26 & 64.41 & 57.89 & 66.73 \\
Gemini 2.5 Pro & 1.50 & 5.00 & 35.82 & 50.33 & 28.52 & 39.44 & 30.00 & 41.39 & 63.03 & 77.32 & 48.01 & 60.92 & 50.98 & 63.62 \\
OpenAI o3-high & 4.50 & 9.00 & 40.07 & 53.13 & 32.25 & 41.74 & 34.00 & 44.07 & 62.79 & 75.39 & 49.76 & 59.25 & 52.61 & 62.27 \\
K2 & 1.12 & 3.50 & 35.12 & 48.82 & 28.12 & 39.51 & 29.67 & 41.35 & 65.12 & 77.28 & 51.34 & 62.24 & 54.36 & 65.08 \\
DeepSeek-R1-0528 & 0.37 & 1.50 & 28.09 & 44.39 & 18.72 & 28.97 & 20.66 & 31.69 & 57.31 & 73.98 & 37.22 & 51.26 & 41.26 & 55.09 \\
Doubao-1.6 & 2.63 & 5.00 & 36.47 & 54.24 & 28.07 & 41.80 & 29.90 & 44.12 & 60.77 & 79.68 & 44.86 & 60.28 & 48.27 & 63.86 \\
Doubao-1.6-non-thinking & 1.00 & 3.50 & 35.16 & 52.07 & 25.29 & 37.44 & 27.21 & 39.86 & 64.63 & 79.62 & 45.18 & 58.44 & 48.97 & 61.98 \\
\bottomrule
\multicolumn{15}{l}{\textit{Multi-Agent Framework on WideSearch-zh}} \\
Claude Sonnet 4 (Thinking) & 2.75 & 6.00 & 41.27 & 58.42 & 35.88 & 50.25 & 36.85 & 51.46 & 64.73 & 79.80 & 55.14 & 67.62 & 57.13 & 69.53 \\
Gemini 2.5 Pro & 1.00 & 4.00 & 36.46 & 50.62 & 29.43 & 40.41 & 30.93 & 42.21 & 62.82 & 73.96 & 48.59 & 58.16 & 51.79 & 60.87 \\
OpenAI o3-high & 2.75 & 6.00 & 41.11 & 59.03 & 32.17 & 45.29 & 33.83 & 47.85 & 62.64 & 79.05 & 47.16 & 59.24 & 50.35 & 63.06 \\
K2 & 1.25 & 3.00 & 40.37 & 56.14 & 33.21 & 46.30 & 34.74 & 48.01 & 67.39 & 78.04 & 53.74 & 64.72 & 56.86 & 66.75 \\
DeepSeek-R1-0528 & 0.50 & 2.00 & 28.67 & 47.38 & 19.37 & 32.63 & 21.17 & 35.08 & 53.66 & 72.11 & 33.79 & 49.38 & 37.66 & 53.15 \\
Doubao-1.6 & 2.25 & 6.00 & 39.22 & 57.11 & 30.84 & 45.21 & 32.83 & 47.49 & 60.34 & 78.62 & 45.51 & 61.32 & 48.79 & 64.43 \\
Doubao-1.6-non-thinking & 0.50 & 1.00 & 36.78 & 56.15 & 24.76 & 37.75 & 26.93 & 40.30 & 64.13 & 79.27 & 42.54 & 56.06 & 46.52 & 59.63 \\
\midrule
\multicolumn{15}{l}{\textit{Multi-Agent Framework on WideSearch-en}} \\
Claude Sonnet 4 (Thinking) & 4.50 & 7.00 & 43.35 & 57.86 & 39.24 & 51.76 & 40.13 & 52.91 & 72.48 & 82.49 & 65.83 & 75.80 & 67.21 & 76.72 \\
Gemini 2.5 Pro & 3.00 & 9.00 & 39.35 & 52.54 & 34.87 & 45.68 & 36.00 & 47.06 & 70.41 & 80.19 & 60.96 & 70.19 & 63.06 & 71.75 \\
OpenAI o3-high & 7.50 & 13.00 & 46.30 & 59.86 & 40.11 & 51.16 & 41.78 & 53.20 & 71.80 & 82.84 & 61.87 & 73.16 & 64.27 & 74.80 \\
K2 & 4.75 & 10.00 & 40.33 & 54.90 & 36.90 & 50.43 & 37.71 & 51.20 & 70.91 & 80.53 & 63.83 & 73.79 & 65.44 & 74.68 \\
DeepSeek-R1-0528 & 1.00 & 4.00 & 29.28 & 45.93 & 23.17 & 36.17 & 24.57 & 38.10 & 60.58 & 77.93 & 48.19 & 65.01 & 50.91 & 67.54 \\
Doubao-1.6 & 2.75 & 5.00 & 39.49 & 57.64 & 33.89 & 48.73 & 35.14 & 50.38 & 68.65 & 83.90 & 58.18 & 73.38 & 60.39 & 74.87 \\
Doubao-1.6-non-thinking & 3.75 & 8.00 & 38.51 & 52.54 & 31.05 & 43.42 & 32.38 & 44.99 & 70.74 & 81.92 & 56.27 & 68.24 & 58.97 & 70.62 \\
\midrule
\multicolumn{15}{l}{\textit{Multi-Agent Framework on WideSearch-all}} \\
Claude Sonnet 4 (Thinking) & 3.62 & 6.50 & 42.31 & 58.14 & 37.56 & 51.01 & 38.49 & 52.19 & 68.60 & 81.15 & 60.48 & 71.71 & 62.17 & 73.13 \\
Gemini 2.5 Pro & 2.00 & 6.50 & 37.90 & 51.58 & 32.15 & 43.04 & 33.47 & 44.64 & 66.62 & 77.07 & 54.77 & 64.18 & 57.42 & 66.31 \\
OpenAI o3-high & 5.12 & 9.50 & 43.70 & 59.44 & 36.14 & 48.23 & 37.80 & 50.52 & 67.22 & 80.95 & 54.51 & 66.20 & 57.31 & 68.93 \\
K2 & 3.00 & 6.50 & 40.35 & 55.52 & 35.06 & 48.36 & 36.22 & 49.60 & 69.15 & 79.28 & 58.78 & 69.26 & 61.15 & 70.72 \\
DeepSeek-R1-0528 & 0.75 & 3.00 & 28.97 & 46.66 & 21.27 & 34.40 & 22.87 & 36.59 & 57.12 & 75.02 & 40.99 & 57.20 & 44.28 & 60.34 \\
Doubao-1.6 & 2.50 & 5.50 & 39.35 & 57.38 & 32.36 & 46.97 & 33.98 & 48.93 & 64.49 & 81.26 & 51.85 & 67.35 & 54.59 & 69.65 \\
Doubao-1.6-non-thinking & 2.12 & 4.50 & 37.64 & 54.34 & 27.91 & 40.58 & 29.65 & 42.65 & 67.43 & 80.59 & 49.41 & 62.15 & 52.75 & 65.13 \\
\bottomrule
\multicolumn{15}{l}{\textit{End-to-End Systems on WideSearch-zh}} \\
Claude & 0.00 & 0.00 & 27.91 & 38.94 & 19.29 & 27.30 & 20.84 & 28.92 & 57.15 & 67.96 & 40.25 & 49.54 & 43.51 & 52.14 \\
Gemini & 1.50 & 4.00 & 37.25 & 47.95 & 31.28 & 39.12 & 32.32 & 40.52 & 64.06 & 73.90 & 50.07 & 58.17 & 52.92 & 60.44 \\
OpenAI o3 & 3.00 & 5.00 & 35.41 & 52.21 & 25.91 & 36.29 & 27.40 & 38.34 & 62.29 & 75.94 & 42.73 & 53.53 & 46.03 & 56.51 \\
\midrule
\multicolumn{15}{l}{\textit{End-to-End Systems on WideSearch-en}} \\
Claude & 5.00 & 10.00 & 30.49 & 43.85 & 26.48 & 36.78 & 27.39 & 38.07 & 60.23 & 74.22 & 51.16 & 62.83 & 53.29 & 64.81 \\
Gemini & 7.00 & 12.00 & 44.43 & 54.58 & 39.75 & 49.13 & 40.95 & 50.29 & 71.39 & 80.43 & 63.30 & 72.19 & 65.18 & 73.90 \\
OpenAI o3 & 3.00 & 6.00 & 24.38 & 40.65 & 19.38 & 31.88 & 20.42 & 33.72 & 55.32 & 67.22 & 42.45 & 53.73 & 45.02 & 56.47 \\
\midrule
\multicolumn{15}{l}{\textit{End-to-End Systems on WideSearch-all}} \\
Claude & 2.50 & 5.00 & 29.20 & 41.40 & 22.89 & 32.04 & 24.11 & 33.49 & 58.69 & 71.09 & 45.70 & 56.18 & 48.40 & 58.47 \\
Gemini & 4.25 & 8.00 & 40.84 & 51.27 & 35.51 & 44.13 & 36.63 & 45.41 & 67.72 & 77.16 & 56.69 & 65.18 & 59.05 & 67.17 \\
OpenAI o3 & 3.00 & 5.50 & 29.90 & 46.43 & 22.64 & 34.08 & 23.91 & 36.03 & 58.81 & 71.58 & 42.59 & 53.63 & 45.52 & 56.49 \\
\bottomrule
\end{tabular}
}
\end{table*}

%% file: paper.bbl
\begin{thebibliography}{32}
\providecommand{\natexlab}[1]{#1}
\providecommand{\url}[1]{\texttt{#1}}
\expandafter\ifx\csname urlstyle\endcsname\relax
  \providecommand{\doi}[1]{doi: #1}\else
  \providecommand{\doi}{doi: \begingroup \urlstyle{rm}\Url}\fi

\bibitem[{Anthropic}(2025)]{Claude4}
{Anthropic}.
\newblock {Introducing Claude 4}.
\newblock \url{https://www.anthropic.com/news/claude-4}, 2025.

\bibitem[Chen et~al.(2025)Chen, Ren, Liu, Hu, Tian, Xie, Liu, Zhang, Liu, Gong, et~al.]{chen2025xbench}
Kaiyuan Chen, Yixin Ren, Yang Liu, Xiaobo Hu, Haotong Tian, Tianbao Xie, Fangfu Liu, Haoye Zhang, Hongzhang Liu, Yuan Gong, et~al.
\newblock xbench: Tracking agents productivity scaling with profession-aligned real-world evaluations.
\newblock \emph{arXiv preprint arXiv:2506.13651}, 2025.

\bibitem[Du et~al.(2025)Du, Xu, Zhu, Wang, and Mao]{du2025deepresearch}
Mingxuan Du, Benfeng Xu, Chiwei Zhu, Xiaorui Wang, and Zhendong Mao.
\newblock Deepresearch bench: A comprehensive benchmark for deep research agents.
\newblock \emph{arXiv preprint arXiv:2506.11763}, 2025.

\bibitem[{Google Gemini}(2025{\natexlab{a}})]{GoogleGeminiDeepResearch2025}
{Google Gemini}.
\newblock {Deep Research is now available on Gemini 2.5 Pro Experimental}.
\newblock \url{https://blog.google/products/gemini/deep-research-gemini-2-5-pro-experimental/}, 2025{\natexlab{a}}.

\bibitem[{Google Gemini}(2025{\natexlab{b}})]{Google_Gemini25Pro}
{Google Gemini}.
\newblock {Gemini 2.5: Our most intelligent AI model}.
\newblock \url{https://blog.google/technology/google-deepmind/gemini-model-thinking-updates-march-2025/}, 2025{\natexlab{b}}.

\bibitem[Guo et~al.(2025)Guo, Yang, Zhang, Song, Zhang, Xu, Zhu, Ma, Wang, Bi, et~al.]{guo2025deepseek}
Daya Guo, Dejian Yang, Haowei Zhang, Junxiao Song, Ruoyu Zhang, Runxin Xu, Qihao Zhu, Shirong Ma, Peiyi Wang, Xiao Bi, et~al.
\newblock Deepseek-r1: Incentivizing reasoning capability in llms via reinforcement learning.
\newblock \emph{arXiv preprint arXiv:2501.12948}, 2025.

\bibitem[Ho et~al.(2020)Ho, Nguyen, Sugawara, and Aizawa]{ho2020constructing}
Xanh Ho, Anh-Khoa~Duong Nguyen, Saku Sugawara, and Akiko Aizawa.
\newblock Constructing a multi-hop qa dataset for comprehensive evaluation of reasoning steps.
\newblock In \emph{Proceedings of the 28th International Conference on Computational Linguistics}, 2020.

\bibitem[Huang et~al.(2025)Huang, Yuan, Ju, Zhao, and Liu]{huang2025reinforced}
Ziyang Huang, Xiaowei Yuan, Yiming Ju, Jun Zhao, and Kang Liu.
\newblock Reinforced internal-external knowledge synergistic reasoning for efficient adaptive search agent.
\newblock \emph{arXiv preprint arXiv:2505.07596}, 2025.

\bibitem[Jin et~al.(2025)Jin, Zeng, Yue, Yoon, Arik, Wang, Zamani, and Han]{jin2025search}
Bowen Jin, Hansi Zeng, Zhenrui Yue, Jinsung Yoon, Sercan Arik, Dong Wang, Hamed Zamani, and Jiawei Han.
\newblock Search-r1: Training llms to reason and leverage search engines with reinforcement learning.
\newblock \emph{arXiv preprint arXiv:2503.09516}, 2025.

\bibitem[Joshi et~al.(2017)Joshi, Choi, Weld, and Zettlemoyer]{joshi2017triviaqa}
Mandar Joshi, Eunsol Choi, Daniel Weld, and Luke Zettlemoyer.
\newblock Triviaqa: A large scale distantly supervised challenge dataset for reading comprehension.
\newblock In \emph{Proceedings of the 55th Annual Meeting of the Association for Computational Linguistics}, 2017.

\bibitem[Kwiatkowski et~al.(2019)Kwiatkowski, Palomaki, Redfield, Collins, Parikh, Alberti, Epstein, Polosukhin, Devlin, Lee, et~al.]{kwiatkowski2019natural}
Tom Kwiatkowski, Jennimaria Palomaki, Olivia Redfield, Michael Collins, Ankur Parikh, Chris Alberti, Danielle Epstein, Illia Polosukhin, Jacob Devlin, Kenton Lee, et~al.
\newblock Natural questions: a benchmark for question answering research.
\newblock \emph{Transactions of the Association for Computational Linguistics}, 7:\penalty0 453--466, 2019.

\bibitem[Lei et~al.(2025)Lei, Chen, Ye, Cao, Shin, SU, SUO, Gao, Hu, Yin, Zhong, Xiong, Sun, Liu, Wang, and Yu]{lei2025spider}
Fangyu Lei, Jixuan Chen, Yuxiao Ye, Ruisheng Cao, Dongchan Shin, Hongjin SU, ZHAOQING SUO, Hongcheng Gao, Wenjing Hu, Pengcheng Yin, Victor Zhong, Caiming Xiong, Ruoxi Sun, Qian Liu, Sida Wang, and Tao Yu.
\newblock Spider 2.0: Evaluating language models on real-world enterprise text-to-{SQL} workflows.
\newblock In \emph{The Thirteenth International Conference on Learning Representations}, 2025.
\newblock URL \url{https://openreview.net/forum?id=XmProj9cPs}.

\bibitem[Li et~al.(2025{\natexlab{a}})Li, Zhang, Yin, Zhang, Ou, Wu, Yin, Li, Tao, Wang, et~al.]{li2025websailor}
Kuan Li, Zhongwang Zhang, Huifeng Yin, Liwen Zhang, Litu Ou, Jialong Wu, Wenbiao Yin, Baixuan Li, Zhengwei Tao, Xinyu Wang, et~al.
\newblock Websailor: Navigating super-human reasoning for web agent.
\newblock \emph{arXiv preprint arXiv:2507.02592}, 2025{\natexlab{a}}.

\bibitem[Li et~al.(2025{\natexlab{b}})Li, Jin, Dong, Qian, Zhu, Wu, Wen, and Dou]{li2025webthinker}
Xiaoxi Li, Jiajie Jin, Guanting Dong, Hongjin Qian, Yutao Zhu, Yongkang Wu, Ji-Rong Wen, and Zhicheng Dou.
\newblock Webthinker: Empowering large reasoning models with deep research capability.
\newblock \emph{arXiv preprint arXiv:2504.21776}, 2025{\natexlab{b}}.

\bibitem[{Manus}(2025)]{Manus2025}
{Manus}.
\newblock {Leave it to Manus}.
\newblock \url{https://manus.im}, 2025.

\bibitem[Mialon et~al.(2023)Mialon, Fourrier, Wolf, LeCun, and Scialom]{mialon2023gaia}
Gr{\'e}goire Mialon, Cl{\'e}mentine Fourrier, Thomas Wolf, Yann LeCun, and Thomas Scialom.
\newblock Gaia: a benchmark for general ai assistants.
\newblock In \emph{The Twelfth International Conference on Learning Representations}, 2023.

\bibitem[{Moonshot AI}(2025)]{KimiResearcher2025}
{Moonshot AI}.
\newblock {Kimi-Researcher: End-to-End RL Training for Emerging Agentic Capabilities}.
\newblock \url{https://moonshotai.github.io/Kimi-Researcher/}, 2025.

\bibitem[{OpenAI}(2024)]{OpenAI_DRSC}
{OpenAI}.
\newblock {Deep Research System Card}.
\newblock \url{https://openai.com/index/deep-research-system-card/}, 2024.

\bibitem[{OpenAI}(2025)]{OpenAI_o3}
{OpenAI}.
\newblock {OpenAI o3 and o4-mini System Card}.
\newblock \url{https://cdn.openai.com/pdf/2221c875-02dc-4789-800b-e7758f3722c1/o3-and-o4-mini-system-card.pdf}, 2025.

\bibitem[Qiu et~al.(2025)Qiu, Qi, Zhang, Juan, Guo, Lu, Wang, Yao, Ren, Jiang, et~al.]{qiu2025alita}
Jiahao Qiu, Xuan Qi, Tongcheng Zhang, Xinzhe Juan, Jiacheng Guo, Yifu Lu, Yimin Wang, Zixin Yao, Qihan Ren, Xun Jiang, et~al.
\newblock Alita: Generalist agent enabling scalable agentic reasoning with minimal predefinition and maximal self-evolution.
\newblock \emph{arXiv preprint arXiv:2505.20286}, 2025.

\bibitem[Seed et~al.(2025)Seed, Chen, Fan, Liu, Liu, Lin, Wang, Wang, Wei, Xu, et~al.]{seed2025seed1}
ByteDance Seed, Jiaze Chen, Tiantian Fan, Xin Liu, Lingjun Liu, Zhiqi Lin, Mingxuan Wang, Chengyi Wang, Xiangpeng Wei, Wenyuan Xu, et~al.
\newblock Seed1. 5-thinking: Advancing superb reasoning models with reinforcement learning.
\newblock \emph{arXiv preprint arXiv:2504.13914}, 2025.

\bibitem[Song et~al.(2025{\natexlab{a}})Song, Jiang, Min, Chen, Chen, Zhao, Fang, and Wen]{song2025r1}
Huatong Song, Jinhao Jiang, Yingqian Min, Jie Chen, Zhipeng Chen, Wayne~Xin Zhao, Lei Fang, and Ji-Rong Wen.
\newblock R1-searcher: Incentivizing the search capability in llms via reinforcement learning.
\newblock \emph{arXiv preprint arXiv:2503.05592}, 2025{\natexlab{a}}.

\bibitem[Song et~al.(2025{\natexlab{b}})Song, Jiang, Tian, Chen, Wu, Zhao, Min, Zhao, Fang, and Wen]{song2025r1++}
Huatong Song, Jinhao Jiang, Wenqing Tian, Zhipeng Chen, Yuhuan Wu, Jiahao Zhao, Yingqian Min, Wayne~Xin Zhao, Lei Fang, and Ji-Rong Wen.
\newblock R1-searcher++: Incentivizing the dynamic knowledge acquisition of llms via reinforcement learning.
\newblock \emph{arXiv preprint arXiv:2505.17005}, 2025{\natexlab{b}}.

\bibitem[Sun et~al.(2025)Sun, Qiao, Guo, Fan, Hou, Jiang, Xie, Zhang, Huang, and Zhou]{sun2025zerosearch}
Hao Sun, Zile Qiao, Jiayan Guo, Xuanbo Fan, Yingyan Hou, Yong Jiang, Pengjun Xie, Yan Zhang, Fei Huang, and Jingren Zhou.
\newblock Zerosearch: Incentivize the search capability of llms without searching.
\newblock \emph{arXiv preprint arXiv:2505.04588}, 2025.

\bibitem[Team et~al.(2025)Team, Bai, Bao, Chen, Chen, Chen, Chen, Chen, Chen, Chen, et~al.]{team2025kimi}
Kimi Team, Yifan Bai, Yiping Bao, Guanduo Chen, Jiahao Chen, Ningxin Chen, Ruijue Chen, Yanru Chen, Yuankun Chen, Yutian Chen, et~al.
\newblock Kimi k2: Open agentic intelligence.
\newblock \emph{arXiv preprint arXiv:2507.20534}, 2025.

\bibitem[Trivedi et~al.(2022)Trivedi, Balasubramanian, Khot, and Sabharwal]{trivedi2022musique}
Harsh Trivedi, Niranjan Balasubramanian, Tushar Khot, and Ashish Sabharwal.
\newblock Musique: Multihop questions via single-hop question composition.
\newblock \emph{Transactions of the Association for Computational Linguistics}, 10:\penalty0 539--554, 2022.

\bibitem[Wei et~al.(2025)Wei, Sun, Papay, McKinney, Han, Fulford, Chung, Passos, Fedus, and Glaese]{wei2025browsecomp}
Jason Wei, Zhiqing Sun, Spencer Papay, Scott McKinney, Jeffrey Han, Isa Fulford, Hyung~Won Chung, Alex~Tachard Passos, William Fedus, and Amelia Glaese.
\newblock Browsecomp: A simple yet challenging benchmark for browsing agents.
\newblock \emph{arXiv preprint arXiv:2504.12516}, 2025.

\bibitem[Wu et~al.(2025)Wu, Li, Fang, Yin, Zhang, Tao, Zhang, Xi, Jiang, Xie, et~al.]{wu2025webdancer}
Jialong Wu, Baixuan Li, Runnan Fang, Wenbiao Yin, Liwen Zhang, Zhengwei Tao, Dingchu Zhang, Zekun Xi, Yong Jiang, Pengjun Xie, et~al.
\newblock Webdancer: Towards autonomous information seeking agency.
\newblock \emph{arXiv preprint arXiv:2505.22648}, 2025.

\bibitem[{x.ai}(2025)]{xAIGrok3Beta2025}
{x.ai}.
\newblock {Grok 3 beta --- the age of reasoning agents}.
\newblock \url{https://x.ai/news/grok-3}, 2025.

\bibitem[Yang et~al.(2018)Yang, Qi, Zhang, Bengio, Cohen, Salakhutdinov, and Manning]{yang2018hotpotqa}
Zhilin Yang, Peng Qi, Saizheng Zhang, Yoshua Bengio, William Cohen, Ruslan Salakhutdinov, and Christopher~D Manning.
\newblock Hotpotqa: A dataset for diverse, explainable multi-hop question answering.
\newblock In \emph{Proceedings of the 2018 Conference on Empirical Methods in Natural Language Processing}, 2018.

\bibitem[Zheng et~al.(2025)Zheng, Fu, Hu, Cai, Ye, Lu, and Liu]{zheng2025deepresearcher}
Yuxiang Zheng, Dayuan Fu, Xiangkun Hu, Xiaojie Cai, Lyumanshan Ye, Pengrui Lu, and Pengfei Liu.
\newblock Deepresearcher: Scaling deep research via reinforcement learning in real-world environments.
\newblock \emph{arXiv preprint arXiv:2504.03160}, 2025.

\bibitem[Zhou et~al.(2025)Zhou, Leon, Ying, Zhang, Shao, Ye, Chong, Jin, Xie, Cao, et~al.]{zhou2025browsecomp}
Peilin Zhou, Bruce Leon, Xiang Ying, Can Zhang, Yifan Shao, Qichen Ye, Dading Chong, Zhiling Jin, Chenxuan Xie, Meng Cao, et~al.
\newblock Browsecomp-zh: Benchmarking web browsing ability of large language models in chinese.
\newblock \emph{arXiv preprint arXiv:2504.19314}, 2025.

\end{thebibliography}
